\documentclass[12pt]{article}

\usepackage{verbatim}
\usepackage[longnamesfirst]{natbib}
\usepackage{amsfonts}
\usepackage{amssymb}
\usepackage{amsmath}
\usepackage{mathrsfs}
\usepackage{rotating}
\usepackage{enumerate}
\usepackage{amsthm}
\usepackage{subfigure}
\usepackage[capposition=top]{floatrow}
\usepackage{url}

\usepackage{booktabs}
\usepackage{colortbl}
\usepackage{array}
\newcolumntype{K}[1]{>{\centering\arraybackslash}p{#1}}

\allowdisplaybreaks

\makeatletter
\def\blfootnote{\xdef\@thefnmark{}\@footnotetext}
\makeatother

\makeatletter
\def\monthname{\ifcase\month\or
January\or February\or March\or April\or May\or June\or
July\or August\or September\or October\or November\or December\fi}
\makeatother

\makeatletter
\renewcommand{\section}{\@startsection{section}{1}{0mm}{-\baselineskip}{0.25\baselineskip}{\centering\normalfont\normalsize\bf}}
\renewcommand{\subsection}{\@startsection{subsection}{2}{0mm}{-\baselineskip}{0.25\baselineskip}{\raggedright\normalfont\normalsize\bf}}
\renewcommand{\subsubsection}{\@startsection{subsubsection}{3}{0mm}{-\baselineskip}{0.25\baselineskip}{\raggedright\normalfont\small\bf}}
\makeatother

\renewcommand{\appendix}{\footnotesize\parindent 0cm\parskip 5pt\setcounter{equation}{0}
\setcounter{section}{0}
\renewcommand{\thesection}{A.\arabic{section}}
\renewcommand{\theequation}{A.\arabic{equation}}
\setcounter{lemma}{0}\renewcommand{\thelemma}{A.\arabic{lemma}}}

\renewenvironment{abstract}
 {\begin{center}\normalsize\bf\text{Abstract}
 \end{center}\begin{quote}\normalsize}
 {\end{quote}}

\newtheoremstyle{break}
  {\topsep}%
  {\topsep}%
  {\itshape}{}%
  {\bfseries}{}%
  {\newline}{}
\theoremstyle{break}
\newtheorem{theorem}{Theorem}
\newtheorem{proposition}{Proposition}

\newtheorem{lemma}{Lemma}
\newtheorem{corollary}{Corollary}

\DeclareMathOperator*{\argmin}{argmin\;}

\DeclareMathOperator{\tr}{tr}

\newcommand{\bs}{\boldsymbol}

\newcommand\quelle[1]{{%
      \unskip\nobreak\hfil\penalty50
      \hskip2em\hbox{}\nobreak\hfil\textbf{#1}%
      \parfillskip=0pt \finalhyphendemerits=0 \par}}

\begin{document}
\vspace*{0.2cm}

\setcounter{page}{0}
\thispagestyle{empty}
\vskip 10pt
\vskip 10pt
\centerline{\Large\bf The Risk of Machine Learning}
  \begin{center}%
    \vskip 10pt
    \blfootnote{\hspace*{-0.25in}Alberto Abadie, Department of Economics, Massachusetts Institute of Technology, abadie@mit.edu. Maximilian Kasy, Department of Economics, Harvard University, maximiliankasy@fas.harvard.edu. We thank Gary Chamberlain, Ellora Derenoncourt, Jiaying Gu, J\'{e}r\'{e}my, L'Hour, Jos\'{e} Luis Montiel Olea, Jann Spiess, Stefan Wager and seminar participants at several institutions for helpful comments and discussions.
}
    {\large
     \lineskip .75em%
      \begin{tabular}[t]{ccc}%
       Alberto Abadie&&Maximilian Kasy\\[1ex]
       MIT&&Harvard University\\
      \end{tabular}
      \par}%
      \vskip 2em%
      {\large \today } \par
       \vskip 2em%
  \end{center}\par

\begin{abstract}
\noindent Many applied settings in empirical economics involve simultaneous estimation of a large number of parameters. In particular, applied economists are often interested in estimating the effects of many-valued treatments (like teacher effects or location effects), treatment effects for many groups, and prediction models with many regressors. In these settings, machine learning methods that combine regularized estimation and data-driven choices of regularization parameters are useful to avoid over-fitting. In this article, we analyze the performance of a class of machine learning estimators that includes ridge, lasso and pretest in contexts that require simultaneous estimation of many parameters. Our analysis aims to provide guidance to applied researchers on (i) the choice between regularized estimators in practice and (ii) data-driven selection of regularization parameters. To address (i), we characterize the risk (mean squared error) of regularized estimators and derive their relative performance as a function of simple features of the data generating process. To address (ii), we show that data-driven choices of regularization parameters, based on Stein's unbiased risk estimate or on cross-validation, yield estimators with risk uniformly close to the risk attained under the optimal (unfeasible) choice of regularization parameters. We use data from recent examples in the empirical economics literature to illustrate the practical applicability of our results.
\end{abstract}



\newpage
\setcounter{page}{1}
\addtolength{\baselineskip}{0.5\baselineskip}

\section{Introduction}
\label{section:introduction}

Applied economists often confront problems that require estimation of a large number of parameters. Examples include (a) estimation of causal (or predictive) effects for a large number of treatments such as neighborhoods or cities, teachers, workers and firms, or judges; (b) estimation of the causal effect of a given treatment for a large number of subgroups; and (c) prediction problems with a large number of predictive covariates or transformations of covariates. The machine learning literature provides a host of estimation methods, such as ridge, lasso, and pretest, which are particularly well adapted to high-dimensional problems. In view of the variety of available methods, the applied researcher faces the question of which of these procedures to adopt in any given situation. This article provides guidance on this choice based on the study of the risk properties (mean squared error) of a class of regularization-based machine learning methods.

A practical concern that generally motivates the adoption of machine learning procedures is the potential for severe over-fitting in high-dimensional settings. To avoid over-fitting, most machine learning procedures for ``supervised learning'' (that is, regression and classification methods for prediction) involve two key features, (i) regularized estimation and (ii) data-driven choice of regularization parameters. These features are also central to more familiar non-parametric estimation methods in econometrics, such as kernel or series regression.

\paragraph{Setup}
In this article, we consider the canonical problem of estimating the unknown means, $\mu_1,\ldots, \mu_n$, of a potentially large set of observed random  variables, $X_1, \ldots, X_n$. After some transformations, our setup covers applications (a)-(c) mentioned above and many others. For example, in the context of a randomized experiment with $n$ subgroups, $X_i$ is the difference in the sample averages of an outcome variable between treated and non-treated for subgroup $i$, and $\mu_i$ is the average treatment effect on the same outcome and subgroup. Moreover, as we discuss in Section \ref{ssec:examples}, the many means problem analyzed in this article encompasses the problem of nonparametric estimation of a regression function.

We consider componentwise estimators of the form $\widehat{\mu}_i=m(X_i, \lambda)$, where $\lambda$ is a non-negative regularization parameter. Typically, $m(x,0)=x$, so that $\lambda=0$ corresponds to the unregularized estimator $\widehat\mu_i=X_i$. Positive values of $\lambda$ typically correspond to regularized estimators, which shrink towards zero, $|\widehat\mu_i|\leq |X_i|$. The value $\lambda=\infty$ typically implies maximal shrinkage: $\widehat\mu_i=0$ for $i=1,\ldots, n$. Shrinkage towards zero is a convenient normalization but it is not essential. Shifting $X_i$ by a constant to $X_i-c$, for $i=1, \ldots, n$, results in shrinkage towards $c$.  

\paragraph{The risk function of regularized estimators}
Our article is structured according to the two mentioned features of machine learning procedures, regularization and data-driven choice of regularization parameters. We first focus on feature (i) and study the risk properties (mean squared error) of regularized estimators with fixed and with oracle-optimal regularization parameters. We show that for any given data generating process there is an (infeasible) risk-optimal regularized componentwise estimator. This estimator has the form of the posterior mean of $\mu_I$ given $X_I$ and given the empirical distribution of $\mu_1, \ldots, \mu_n$, where $I$ is a random variable with uniform distribution on the set of indices $\{1,2,\ldots, n\}$. The optimal regularized estimator is useful to characterize the risk properties of machine learning estimators. It turns out that, in our setting, the risk function of any regularized estimator can be expressed as a function of the distance between that regularized estimator and the optimal one.

Instead of conditioning on $\mu_1, \ldots, \mu_n$, one can consider the case where each $(X_i,\mu_i)$ is a realization of a random vector $(X,\mu)$ with distribution $\pi$ and a notion of risk that is integrated over the distribution of $\mu$ in the population. For this alternative definition of risk, we derive results analogous to those of the previous paragraph.

We next turn to a family of parametric models for $\pi$. We consider models that allow for a probability mass at zero in the distribution of $\mu$, corresponding to the notion of sparsity, while conditional on $\mu\neq 0$ the distribution of $\mu$ is normal around some grand mean. For these parametric models we derive analytic risk functions under oracle choices of risk minimizing values for $\lambda$, which allow for an intuitive discussion of the relative performance of alternative estimators. We focus our attention on three estimators that are widespread in the empirical machine learning literature: ridge, lasso, and pretest.
When the point-mass of true zeros is small, ridge tends to perform better than lasso or pretest. When there is a sizable share of true zeros, the ranking of the estimators depends on the other characteristics of the distribution of $\mu$: (a) if the non-zero parameters are smoothly distributed in a vicinity of zero, ridge still performs best; (b) if most of the distribution of non-zero parameters assigns large probability to a set well-separated from zero, pretest estimation tends to perform well; and (c) lasso tends to do comparatively well in intermediate cases that fall somewhere between (a) and (b), and overall is remarkably robust across the different specifications. This characterization of the relative performance of ridge, lasso, and pretest is consistent with the results that we obtain for the empirical applications discussed later in the article.

\paragraph{Data-driven choice of regularization parameters}
The second part the article turns to feature (ii) of machine learning estimators and studies the data-driven choice of regularization parameters. We consider choices of regularization parameters based on the minimization of a criterion function that estimates risk. Ideally, a machine learning estimator evaluated at a data-driven choice of the regularization parameter would have a risk function that is uniformly close to the risk function of the infeasible estimator using an oracle-optimal regularization parameter (which minimizes true risk). We show this type of uniform consistency can be achieved under fairly mild conditions whenever the dimension of the problem under consideration is large. This is in stark contrast to well-known results in \cite{leeb2006risk} for low-dimensional settings. We further provide fairly weak conditions under which machine learning estimators with data-driven choices of the regularization parameter, based on Stein's unbiased risk estimate (SURE) and on cross-validation (CV), attain uniform risk consistency. In addition to allowing data-driven selection of regularization parameters, uniformly consistent estimation of the risk of shrinkage estimators can be used to select among alternative shrinkage estimators on the basis of their estimated risk in specific empirical settings.

\paragraph{Applications}
We illustrate our results in the context of three applications taken from the empirical economics literature. The first application uses data from \cite{chetty2015impacts} to study the effects of locations on intergenerational earnings mobility of children. The second application uses data from the event-study analysis in \cite{dellavigna2010detecting} who investigate whether the stock prices of weapon-producing companies react to changes in the intensity of conflicts in countries under arms trade embargoes. The third application considers nonparametric estimation of a Mincer equation using data from the Current Population Survey (CPS), as in \cite{belloni2011high}. The presence of many neighborhoods in the first application, many weapon producing companies in the second one, and many series regression terms in the third one makes these estimation problems high-dimensional.

These examples showcase how simple features of the data generating process affect the relative performance of machine learning estimators.
They also illustrate the way in which consistent estimation of the risk of shrinkage estimators can be used to choose regularization parameters and to select among different estimators in practice. For the estimation of location effects in \cite{chetty2015impacts} we find estimates that are not overly dispersed around their mean and no evidence of sparsity. In this setting, ridge outperforms lasso and pretest in terms of estimated mean squared error. In the setting of the event-study analysis in \cite{dellavigna2010detecting}, our results suggest that a large fraction of values of parameters are closely concentrated around zero, while a smaller but non-negligible fraction of parameters are positive and substantially separated from zero. In this setting, pretest dominates. Similarly to the result for the setting in \cite{dellavigna2010detecting}, the estimation of the parameters of a Mincer equation in \cite{belloni2011high} suggests a sparse approximation to the distribution of parameters. Substantial shrinkage at the tails of the distribution is still helpful in this setting, so that lasso dominates.

\paragraph{Roadmap}
The rest of this article is structured as follows. Section \ref{sec:setup} introduces our setup: the canonical problem of estimating a vector of means under quadratic loss. Section \ref{ssec:examples} discusses a series of examples from empirical economics that are covered by our setup.
Section \ref{ssec:literature} discusses the setup of this article in the context of the machine learning literature and of the older literature on estimation of normal means. Section \ref{sec:riskfunction} provides characterizations of the risk function of regularized estimators in our setting. We derive a general characterization in Section \ref{ssec:generalriskfunction}. Sections \ref{ssec:analyticriskfunction} and \ref{ssec:spikeandnormal} provide analytic formulas for risk under additional assumptions. In particular, in Section \ref{ssec:spikeandnormal} we derive analytic formulas for risk in a spike-and-normal model . These characterizations allow for a comparison of the mean squared error of alternative procedures and yield recommendations for the choice of an estimator. Section \ref{sec:datadrivenregularization} turns to data-driven choices of regularization parameters. We show uniform risk consistency results for Stein's unbiased risk estimate and for cross-validation. Section \ref{sec:discussion} discusses extensions and explains the apparent contradiction between our results and those in \cite{leeb2005model}. Section \ref{section:simulations} reports simulation results. 
Section \ref{sec:applications} discusses several empirical applications. Section \ref{sec:conclusion} concludes. The appendix contains proofs and supplemental materials.

\section{Setup}
\label{sec:setup}

Throughout this paper, we consider the following setting. We observe a realization of an $n$-vector of real-valued random variables, $\bs X=(X_1,\ldots , X_n)'$, where the components of $\bs X$ are mutually independent with finite mean $\mu_i$ and finite variance $\sigma_i^2$, for $i=1,\ldots, n$. Our goal is to estimate $\mu_1, \ldots, \mu_n$.

In many applications, the $X_i$ arise as preliminary least squares estimates of the coefficients of interest, $\mu_i$.
Consider, for instance, a randomized controlled trial where randomization of treatment assignment is carried out separately for $n$ non-overlapping subgroups. Within each subgroup, the difference in the sample averages between treated and control units, $X_i$, has mean equal to the average treatment effect for that group in the population, $\mu_i$. Further examples are discussed in Section \ref{ssec:examples} below.

\paragraph{Componentwise estimators}
We restrict our attention to componentwise estimators of $\mu_i$,
\begin{equation*}
\widehat \mu_i = m(X_i,\lambda),
\end{equation*}
where $m:\mathbb R\times [0,\infty]\mapsto \mathbb R$ defines an estimator of $\mu_i$ as a function of $X_i$ and a non-negative regularization parameter, $\lambda$. The parameter $\lambda$ is common across the components $i$ but might depend on the vector $\bs X$. We study data-driven choices $\widehat\lambda$ in Section \ref{sec:datadrivenregularization} below, focusing in particular on Stein's unbiased risk estimate (SURE) and cross-validation (CV).

\begin{figure}
\begin{center}
\caption{{Estimators}}
\label{fig:componetwisem}
\label{fig:mlambda}
\includegraphics[width = \textwidth]{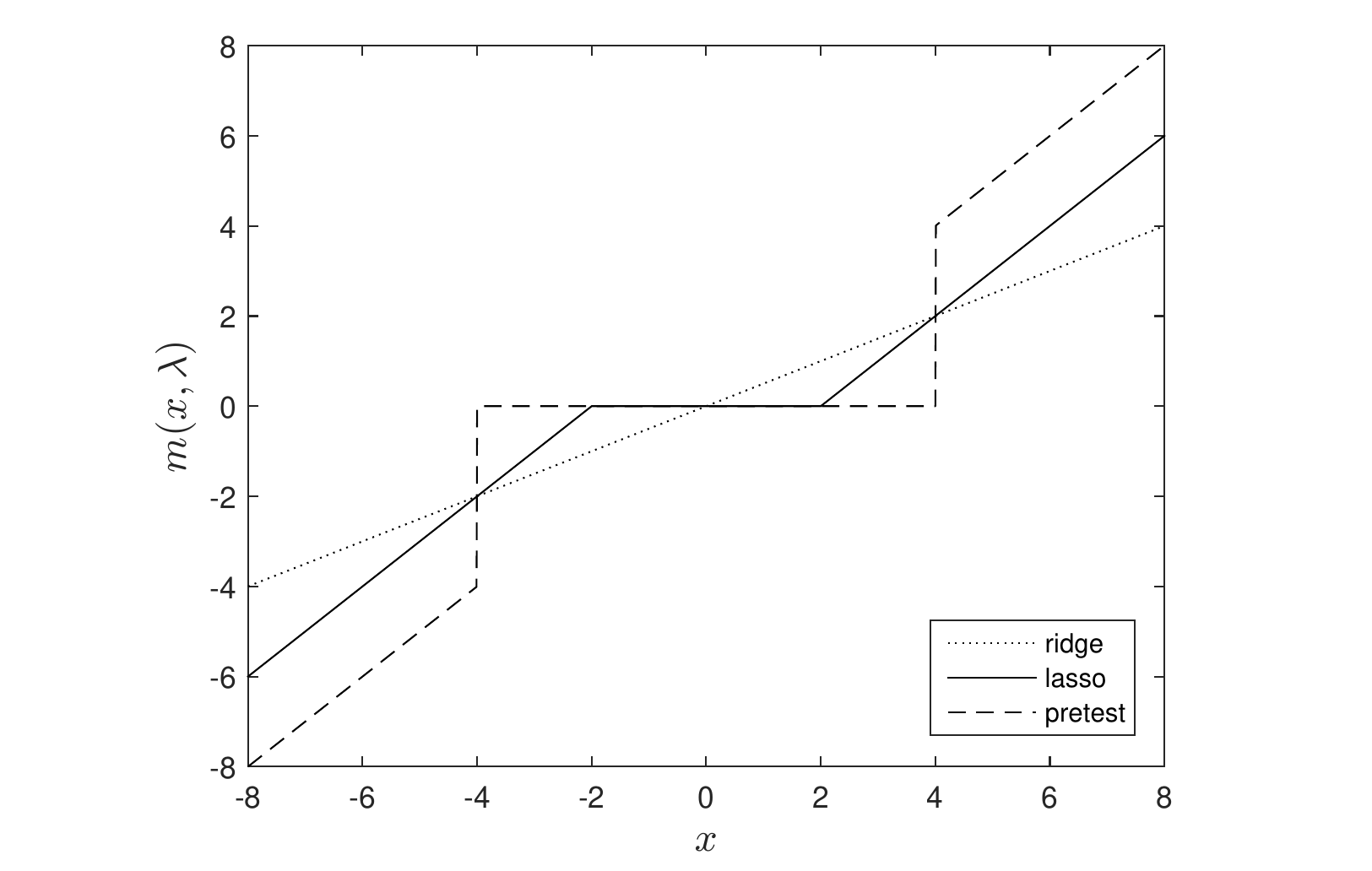}
\end{center}
\floatfoot{\footnotesize This graph plots $m_R(x, \lambda)$, $m_L(x, \lambda)$, and $m_{PT}(x, \lambda)$ as functions of $x$.
 The regularization parameters are $\lambda=1$ for ridge, $\lambda=2$
for lasso, and $\lambda=4$ for pretest.}
\end{figure}

Popular estimators of this componentwise form are ridge, lasso, and pretest.
They are defined as follows:
\begin{align*}
m_R(x,\lambda) &= \argmin_{m\in\mathbb R}\ (x-m)^2+\lambda m^2 &\textrm{(ridge)}\\
&=\frac{1}{1+\lambda}\, x,\\[2ex]
m_L(x,\lambda) &= \argmin_{m\in\mathbb R}\ (x-m)^2+2\lambda |m| &\textrm{(lasso)}\\
&=1(x<-\lambda) (x +\lambda) + 1(x>\lambda) (x -\lambda),\\[2ex]
m_{PT}(x,\lambda) &= \argmin_{m\in\mathbb R}\ (x-m)^2+\lambda^2 1(m\neq 0) &\textrm{(pretest)}\\
&= 1(|x|>\lambda) x,
\end{align*}
where $1(A)$ denotes the indicator function, which equals $1$ if $A$ holds and $0$ otherwise.
Figure \ref{fig:mlambda} plots $m_R(x, \lambda)$, $m_L(x, \lambda)$ and $m_{PT}(x, \lambda)$ as functions of $x$. For reasons apparent in Figure \ref{fig:mlambda}, ridge, lasso, and pretest estimators are sometimes referred to as linear shrinkage, soft thresholding, and hard thresholding, respectively. As we discuss below, the problem of determining the optimal choice among these estimators in terms of minimizing mean squared error is equivalent to the problem of determining which of these estimators best approximates a certain optimal estimating function, $m^*$.

Let $\bs\mu=(\mu_1,\ldots , \mu_n)'$ and $\widehat{\bs\mu}=(\widehat\mu_1,\ldots , \widehat\mu_n)'$, where for simplicity we leave the dependence of $\widehat\mu$ on $\lambda$ implicit in our notation. Let $P_1, \ldots, P_n$ be the distributions of $X_1,\ldots, X_n$, and let $\bs P=(P_1,\ldots, P_n)$.

\paragraph{Loss and risk}
We evaluate estimates based on the squared error loss function, or compound loss,
\begin{equation*}
L_n(\bs X,m(\cdot,\lambda),\bs P) =  \frac{1}{n} \sum_{i=1}^n \big(m(X_i,\lambda)- \mu_i\big)^2,
\end{equation*}
where $L_n$ depends on $\bs P$ via $\bs \mu$. We will use expected loss to rank estimators. There are different ways of taking this expectation, resulting in different risk functions, and the distinction between them is conceptually important.

\emph{Componentwise risk} fixes $P_i$ and considers the expected squared error of $\widehat{\mu}_i$ as an estimator of $\mu_i$,
\begin{equation*}
R(m(\cdot,\lambda),P_i)=E[(m(X_i,\lambda)-\mu_i)^2|P_i].
\end{equation*}

{\em Compound risk} averages componentwise risk over the empirical distribution of $P_i$ across the components $i=i,\ldots,n$. Compound risk is given by the expectation of compound loss $L_n$ given $\bs P$,
\begin{align*}
R_n(m(\cdot,\lambda),\bs P) &= E[L_n({\bs X},m(\cdot,\lambda),\bs P)|\bs P] \nonumber\\
& =\frac{1}{n}\sum_{i=1}^n E[(m(X_i,\lambda)-\mu_i)^2|P_i] \nonumber\\
& =\frac{1}{n}\sum_{i=1}^n R(m(\cdot,\lambda),P_i).
\end{align*}

Finally, {\em integrated} (or {\em empirical Bayes}) {\em risk} considers $P_1, \ldots, P_n$ to be themselves draws from some population distribution, $\Pi$. This induces a joint distribution, $\pi$, for $(X_i,\mu_i)$. Throughout the article, we will often use a subscript $\pi$ to denote characteristics of the joint distribution of $(X_i,\mu_i)$. Integrated risk refers to
loss integrated over $\pi$ or, equivalently, componentwise risk integrated over $\Pi$,
\begin{align}
\label{equation:intrisk}
\bar R(m(\cdot,\lambda), \pi) & = E_\pi[L_n({\bs X},m(\cdot,\lambda),\bs P)]\nonumber\\
&= E_\pi[(m(X_i,\lambda)-\mu_i)^2] \nonumber\\
&=\int R(m(\cdot,\lambda),P_i) d\Pi(P_i).
\end{align}
Notice the similarity between compound risk and integrated risk: they differ only by replacing an empirical (sample) distribution by a population distribution. For large $n$, the difference between the two vanishes, as we will explore in Section \ref{sec:datadrivenregularization}.

\paragraph{Regularization parameter}
Throughout, we will use  $R_n(m(\cdot,\lambda), \bs P)$  to denote the risk function of the estimator $m(\cdot,\lambda)$ with fixed (non-random) $\lambda$, and similarly for $\bar R(m(\cdot,\lambda), \pi)$. In contrast, $R_n(m(\cdot,\widehat \lambda_n), \bs P)$ is the risk function taking into account the randomness of $\widehat{\lambda}_n$, where the latter is chosen in a data-dependent manner, and similarly for $\bar R(m(\cdot,\widehat \lambda_n), \pi)$.

For a given $\bs P$, we define the ``oracle'' selector of the regularization parameter as the value of $\lambda$ that minimizes compound risk,
\begin{equation*}
\lambda^*(\bs P) = \argmin_{\lambda\in [0,\infty]} R_n(m(\cdot,\lambda),\bs P),
\end{equation*}
whenever the argmin exists.
We use $\lambda^*_R(\bs P)$, $\lambda^*_L(\bs P)$ and $\lambda^*_{PT}(\bs P)$ to denote the oracle selectors for ridge, lasso, and pretest, respectively. Analogously, for a given $\pi$, we define
\begin{equation}
\label{equation:intoracle}
\bar\lambda^*(\pi) = \argmin_{\lambda\in [0,\infty]}\bar R(m(\cdot,\lambda),\pi)
\end{equation}
whenever the argmin exists,
with $\bar\lambda^*_R(\pi)$, $\bar\lambda^*_L(\pi)$, and $\bar\lambda^*_{PT}(\pi)$ for ridge, lasso, and pretest, respectively.
In Section \ref{sec:riskfunction}, we characterize compound and integrated risk for fixed $\lambda$ and for the oracle-optimal $\lambda$.
In Section \ref{sec:datadrivenregularization} we show that data-driven choices $\widehat{\lambda}_n$ are, under certain conditions, as good as the oracle-optimal choice, in a sense to be made precise.

\subsection{Empirical examples}
\label{ssec:examples}

Our setup describes a variety of settings often encountered in empirical economics, where $X_1, \ldots, X_n$ are unbiased or close-to-unbiased but noisy least squares estimates of a set of parameters of interest, $\mu_1,\ldots, \mu_n$. As mentioned in the introduction, examples include (a) studies estimating causal or predictive effects for a large number of treatments such as neighborhoods, cities, teachers, workers, firms, or judges; (b) studies estimating the causal effect of a given treatment for a large number of subgroups; and (c) prediction problems with a large number of predictive covariates or transformations of covariates.

\paragraph{Large number of treatments}
Examples in the first category include \cite{chetty2015impacts}, who estimate the effect of geographic locations on intergenerational mobility for a large number of locations. \citeauthor{chetty2015impacts} use differences between the outcomes of siblings whose parents move during their childhood in order to identify these effects. The problem of estimating a large number of parameters also arises in the teacher value-added literature  when the objects of interest are individual teachers' effects, see, for instance, \cite{chetty2014measuring}. In labor economics, estimation of firm and worker effects in studies of wage inequality has been considered in \cite{abowd1999high}.
Another example within the first category is provided by  \cite{abrams2012judges}, who estimate differences in the effects of defendant's race on sentencing across individual judges.

\paragraph{Treatment for large number of subgroups}
Within the second category, which consists of estimating the effect of a treatment for many sub-populations, our setup can be applied to the estimation of heterogeneous causal effects of class size on student outcomes across many subgroups. For instance, project STAR (\citeauthor{krueger1999experimental}, \citeyear{krueger1999experimental}) involved experimental assignment of students to classes of different sizes in 79 schools. Causal effects for many subgroups are also of interest in medical contexts or for active labor market programs, where doctors / policy makers have to decide on treatment assignment based on individual characteristics. In some empirical settings, treatment impacts are individually estimated for each sample unit. This is often the case in empirical finance, where event studies are used to estimate reactions of stock market prices to newly available information. For example, \cite{dellavigna2010detecting} estimate the effects of changes in the intensity of armed conflicts in countries under arms trade embargoes on the stock market prices of arms-manufacturing companies.

\paragraph{Prediction with many regressors}
The third category is prediction with many regressors. This category fits in the setting of this article after orthogonalization of the regressors. Prediction with many regressors arises, in particular, in macroeconomic forecasting. \cite{stock2012generalized}, in an analysis complementing the present article, evaluate various procedures in terms of their forecast performance for a number of macroeconomic time series for the United States. Regression with many predictors also arises in series regression, where series terms are transformations of a set of predictors. Series regression and its asymptotic properties have been widely studied in econometrics (see for instance  \citeauthor{newey1997convergence}, \citeyear{newey1997convergence}). \citeauthor{wasserman2006all} (\citeyear{wasserman2006all}, Sections 7.2-7.3) provides an illuminating discussion of the equivalence between the normal means model studied in this article and nonparametric regression estimation. For that setting, $X_1,\ldots, X_n$ and $\mu_1,\ldots, \mu_n$ correspond to the estimated and true regression coefficients on an orthogonal basis of functions. Application of lasso and pretesting to series regression is discussed, for instance, in \cite{belloni2011high}. Appendix \ref{asection:prediction} further discusses the relationship between the normal means model and prediction models.

In Section \ref{sec:applications}, we return to three of these applications, revisiting the estimation of location effects on intergenerational mobility, as in \cite{chetty2015impacts}, the effect of changes in the intensity of conflicts in arms-embargo countries on the stock prices of arms manufacturers, as in \cite{dellavigna2010detecting}, and nonparametric series estimation of a Mincer equation, as in \cite{belloni2011high}.

\subsection{Statistical literature}
\label{ssec:literature}

Machine learning methods are becoming widespread in econometrics -- see, for instance, \cite{imbensatheySI2015} and \cite{kleinberg2015prediction}. A large number of estimation procedures are available to the applied researcher. Textbooks such as \cite{friedman2009elements} or \cite{murphy2012machine} provide an introduction to machine learning. Lasso, which was first introduced by \cite{tibshirani1996regression}, is becoming particularly popular in applied economics. \cite{belloni2011high} provide a review of lasso including theoretical results and applications in economics.

Much of the research on machine learning focuses on algorithms and computational issues, while the formal statistical properties of machine learning estimators have received less attention. However, an older and superficially unrelated literature in mathematical statistics and statistical decision theory on the estimation of the normal means model has produced many deep results which turn out to be relevant for understanding the behavior of estimation procedures in non-parametric statistics and machine learning. A foundational article in this literature is \cite{james1961estimation}, who study the case $X_i\sim N(\mu_i, 1)$. They show that the estimator $\widehat{\bs\mu}=\bs X$ is inadmissible whenever $n\geq 3$. That is, there exists a (shrinkage) estimator that has mean squared error smaller than the mean squared error of $\widehat{\bs\mu}=\bs X$ for all values of $\bs \mu$. \cite{brown1971admissible} provides more general characterizations of admissibility and shows that this dependence on dimension is deeply connected to the recurrence or transience of Brownian motion. \cite{stein1981estimation} characterizes the risk function of arbitrary estimators, $\widehat{\bs\mu}$, and based on this characterization proposes an unbiased estimator of the mean squared error of a given estimator, labeled ``Stein's unbiased risk estimator'' or SURE. We return to SURE in Section \ref{section:SURE} as a method to produce data-driven choices of regularization parameters. In section \ref{ssec:CV}, we discuss cross-validation as an alternative method to obtain data-driven choices of regularization parameters in the context studied in this article.\footnote{See, e.g., \cite{arlot2010survey} for a survey on cross-validation methods for model selection.}

A general approach for the construction of regularized estimators, such as the one proposed by \cite{james1961estimation}, is provided by the empirical Bayes framework, first proposed in \cite{robbins1956empirical} and \cite{robbins1964empirical}. A key insight of the empirical Bayes framework, and the closely related compound decision problem framework, is that trying to minimize squared error in higher dimensions involves a trade-off across components of the estimand. The data are informative about which estimators and regularization parameters perform well in terms of squared error and thus allow one to construct regularized estimators that dominate the unregularized  $\widehat{\bs\mu}=\bs X$. This intuition is elaborated on in \cite{stigler1990}. The empirical Bayes framework was developed further by \cite{efron1973stein} and \cite{Morris1983}, among others. Good reviews and introductions can be found in \cite{zhang2003compound} and \cite{efron2010large}.

In Section \ref{sec:datadrivenregularization} we consider data-driven choices of regularization parameters and emphasize uniform validity of asymptotic approximations to the risk function of the resulting estimators. Lack of uniform validity of standard asymptotic characterizations of risk (as well as of test size) in the context of pretest and model-selection based estimators in low-dimensional settings has been emphasized by \cite{leeb2005model}.

While in this article we study risk-optimal estimation of $\bs\mu$, a related literature has focused on the estimation of confidence sets for the same parameter. \citeauthor{wasserman2006all} (\citeyear{wasserman2006all}, Section 7.8) and \cite{casella2012confidence} surveys some results in this literature. \cite{efron2010large} studies hypotheses testing in high dimensional settings from an empirical Bayes perspective.

\section{The risk function}
\label{sec:riskfunction}

We now turn to our first set of formal results, which pertain to the mean squared error of regularized estimators. Our goal is to guide the researcher's choice of estimator by describing the conditions under which each of the alternative machine learning estimators performs better than the others.

We first derive a general characterization of the mean squared error of regularized estimators. This characterization is based on the geometry of estimating functions $m$ as depicted in Figure \ref{fig:mlambda}. It is a-priori not obvious which of these functions is best suited for estimation. We show that for any given data generating process there is an \textit{optimal} function $m^*_{\bs P}$ that minimizes mean squared error. Moreover, we show that the mean squared error for an \textit{arbitrary} $m$ is equal, up to a constant, to the $L^2$ distance between $m$ and $m^*_{\bs P}$. A function $m$ thus yields a good estimator if it is able to approximate the shape of $m^*_{\bs P}$ well.

In Section \ref{ssec:analyticriskfunction}, we provide analytic expressions for the componentwise risk of ridge, lasso, and pretest estimators, imposing the additional assumption of normality. Summing or integrating componentwise risk over some distribution for $(\mu_i, \sigma_i)$ delivers expressions for compound and integrated risk.

In Section \ref{ssec:spikeandnormal}, we turn to a specific parametric family of data generating processes where each $\mu_i$ is equal to zero with probability $p$, reflecting the notion of sparsity, and is otherwise drawn from a normal distribution with some mean $\mu_0$ and variance $\sigma^2_0$. For this parametric family indexed by $(p,\mu_0,\sigma_0)$, we provide analytic risk functions and visual comparisons of the relative performance of alternative estimators. This allows us to identify key features of the data generating process which affect the relative performance of alternative estimators.

\subsection{General characterization}
\label{ssec:generalriskfunction}

Recall the setup introduced in Section \ref{sec:setup}, where we observe $n$ jointly independent random variables $X_1,\ldots, X_n$, with means
$\mu_1,\ldots , \mu_n$. We are interested in the mean squared error for the compound problem of estimating all $\mu_1,\ldots, \mu_n$ simultaneously.
In this formulation of the problem, $\mu_1,\ldots, \mu_n$ are fixed unknown parameters.

Let $I$ be a random variable with a uniform distribution over the set $\{1,2, \ldots, n\}$ and consider the random component $(X_I, \mu_I)$ of $(\bs X, \bs \mu)$. This construction induces a mixture distribution for $(X_I, \mu_I)$ (conditional on $\bs P$),
\[
(X_I, \mu_I)|\bs P \sim \frac{1}{n} \sum_{i=1}^n P_i \delta_{\mu_i},
\]
where $\delta_{\mu_1}, \ldots, \delta_{\mu_n}$ are Dirac measures at $\mu_1,\ldots, \mu_n$.
Based on this mixture distribution, define the conditional expectation
\begin{equation*}
m^*_{\bs P}(x) = E[\mu_I | X_I=x, \bs P]
\end{equation*}
and the average conditional variance
\[v^*_{\bs P} =E\big[\mbox{var}(\mu_I | X_I, \bs P)|\bs P\big]. \]
The next theorem characterizes the compound risk of an estimator in terms of the average squared discrepancy relative to $m_{\bs P}^*$, which implies that $m_{\bs P}^*$ is optimal (lowest mean squared error) for the compound problem.

\begin{theorem}[Characterization of risk functions]
\label{theo:mstar}
Under the assumptions of Section \ref{sec:setup} and $\sup_{\lambda\in[0,\infty]}E[(m(X_I,\lambda))^2|{\bs P}]<\infty$, the compound risk function $R_n$ of $\widehat{\mu}_i=m(X_i,\lambda)$ can be written as
\begin{equation*}
R_n(m(\cdot,\lambda),\bs P) = v^*_{\bs P} + E\big[(m(X_I,\lambda) - m^*_{\bs P}(X_I))^2|\bs P\big],
\end{equation*}
which implies
\begin{equation*}
\lambda^*(\bs P)=
\argmin_{\lambda\in[0,\infty]} E\big[(m(X_I,\lambda) - m^*_{\bs P}(X_I))^2|\bs P\big]
\end{equation*}
whenever $\lambda^*(\bs P)$ is well defined.
\end{theorem}

The proof of this theorem and all further results can be found in the appendix.

The statement of this theorem implies that the risk of componentwise estimators is equal to an irreducible part $v^*_{\bs P}$, plus the $L^2$ distance of the estimating function $m(.,\lambda)$ to the infeasible optimal estimating function $m^*_{\bs P}$. A given data generating process $\bs P$ maps into an optimal estimating function $m^*_{\bs P}$, and the relative performance of alternative estimators $m$ depends on how well they approximate $m^*_{\bs P}$.

We can easily write $m^*_{\bs P}$ explicitly because the conditional expectation defining $m^*_{\bs P}$ is a weighted average of the values taken by $\mu_i$. Suppose, for example, that $X_i\sim N(\mu_i, 1)$ for $i=1\ldots n$. Let $\phi$ be the standard normal probability density function. Then,
\[
m^*_{\bs P}(x) = \frac{\displaystyle\sum_{i=1}^n \mu_i\, \phi(x - \mu_i)}{\displaystyle\sum_{i=1}^n \phi(x - \mu_i)}.
\]

Theorem \ref{theo:mstar} conditions on the empirical distribution of $\mu_1,\ldots, \mu_n$, which corresponds to the notion of compound risk.
Replacing this empirical distribution by the population distribution $\pi$, so that
\[(X_i,\mu_i) \sim \pi,\]
results analogous to those in Theorem \ref{theo:mstar} are obtained for the integrated risk and the integrated oracle selectors in equations (\ref{equation:intrisk}) and (\ref{equation:intoracle}). That is, let
\[
\bar m^*_\pi(x)= E_\pi[\mu_i|X_i=x]
\]
and
\[
\bar v^*_\pi = E_\pi[\mbox{var}_\pi(\mu_i|X_i)],
\]
and assume $\sup_{\lambda\in [0,\infty]} E_\pi[(m(X_i,\lambda)-\mu_i)^2]<\infty$. Then
\[
\bar R(m(\cdot,\lambda),\pi)=\bar v^*_\pi + E_\pi\big[(m(X_i,\lambda) - \bar m^*_{\pi}(X_i))^2\big]
\]
and
\begin{equation}
\label{equation:lsof}
\bar\lambda^*(\pi) = \argmin_{\lambda\in[0,\infty]} E_\pi\big[(m(X_i,\lambda) - \bar m^*_{\pi}(X_i))^2\big].
\end{equation}
The proof of these assertions is analogous to the proof of Theorem \ref{theo:mstar}. $m^*_{\bs P}$ and $\bar m^*_{\pi}$ are optimal componentwise estimators or ``shrinkage functions'' in the sense that they minimize the compound and integrated risk, respectively.

\subsection{Componentwise risk}
\label{ssec:analyticriskfunction}

The characterization of the risk of componentwise estimators in the previous section relies only on the existence of second moments. Explicit expressions for compound risk and integrated risk can be derived under additional structure. We shall now consider a setting in which the $X_i$ are normally distributed,
\[
X_i \sim N(\mu_i, \sigma_i^2).
\]
This is a particularly relevant scenario in applied research, where the $X_i$ are often unbiased estimators with a normal distribution in large samples (as in examples (a) to (c) in Sections \ref{section:introduction} and \ref{ssec:examples}). For concreteness, we will focus on the three widely used componentwise estimators introduced in Section \ref{sec:setup}, ridge, lasso, and pretest, whose estimating functions $m$ were plotted in Figure \ref{fig:componetwisem}. The following lemma provides explicit expressions for the componentwise risk of these estimators.
\begin{lemma}[Componentwise risk]
\label{lem:componentwise}
Consider the setup of Section \ref{sec:setup}. Then, for $i=1,\ldots, n$, the componentwise risk of ridge is:
\[
R(m_R(\cdot,\lambda),P_i) = \left (\frac{1}{1+ \lambda}\right )^2  \sigma^2_i +\left (1 - \frac{1}{1+ \lambda }\right )^2\mu_i^2.
\]
Assume in addition that $X_i$ has a normal distribution. Then, the componentwise risk of lasso is
\begin{align*}
R(m_L(\cdot,\lambda),P_i)&=\Bigg(1 + \Phi\Big(\displaystyle\frac{-\lambda-\mu_i}{\sigma_i}\Big)-\Phi\Big(\displaystyle\frac{\lambda-\mu_i}{\sigma_i}\Big)\Bigg)(\sigma_i^2+\lambda^2)\\
 &+\Bigg(\Big(\displaystyle\frac{-\lambda-\mu_i}{\sigma_i}\Big)\phi\Big(\displaystyle\frac{\lambda-\mu_i}{\sigma_i}\Big)+\Big(\displaystyle\frac{-\lambda+\mu_i}{\sigma_i}\Big)\phi\Big(\displaystyle\frac{-\lambda-\mu_i}{\sigma_i}\Big)\Bigg)\sigma_i^2\\
&+\left(\Phi\Big(\displaystyle\frac{\lambda-\mu_i}{\sigma_i}\Big)-\Phi\Big(\displaystyle\frac{-\lambda-\mu_i}{\sigma_i}\Big)\right)\mu_i^2.
\end{align*}
Under the same conditions, the componentwise risk of pretest is
\begin{align*}
R(m_{PT}(\cdot,\lambda),P_i)&=\Bigg(1 + \Phi\Big(\displaystyle\frac{-\lambda-\mu_i}{\sigma_i}\Big)-\Phi\Big(\displaystyle\frac{\lambda-\mu_i}{\sigma_i}\Big) \Bigg)\sigma_i^2\\ &+\Bigg(\Big(\displaystyle\frac{\lambda-\mu_i}{\sigma_i}\Big)\phi\Big(\displaystyle\frac{\lambda-\mu_i}{\sigma_i}\Big)-\Big(\displaystyle\frac{-\lambda-\mu_i}{\sigma_i}\Big)\phi\Big(\displaystyle\frac{-\lambda-\mu_i}{\sigma_i}\Big)\Bigg)\sigma_i^2\\
&+\left(\Phi\Big(\displaystyle\frac{\lambda-\mu_i}{\sigma_i}\Big)-\Phi\Big(\displaystyle\frac{-\lambda-\mu_i}{\sigma_i}\Big)\right)\mu_i^2.
\end{align*}
\end{lemma}

\begin{figure}\begin{center}
\caption{{Componentwise risk functions}}
\label{fig:comprisk}
\includegraphics[width = \textwidth]{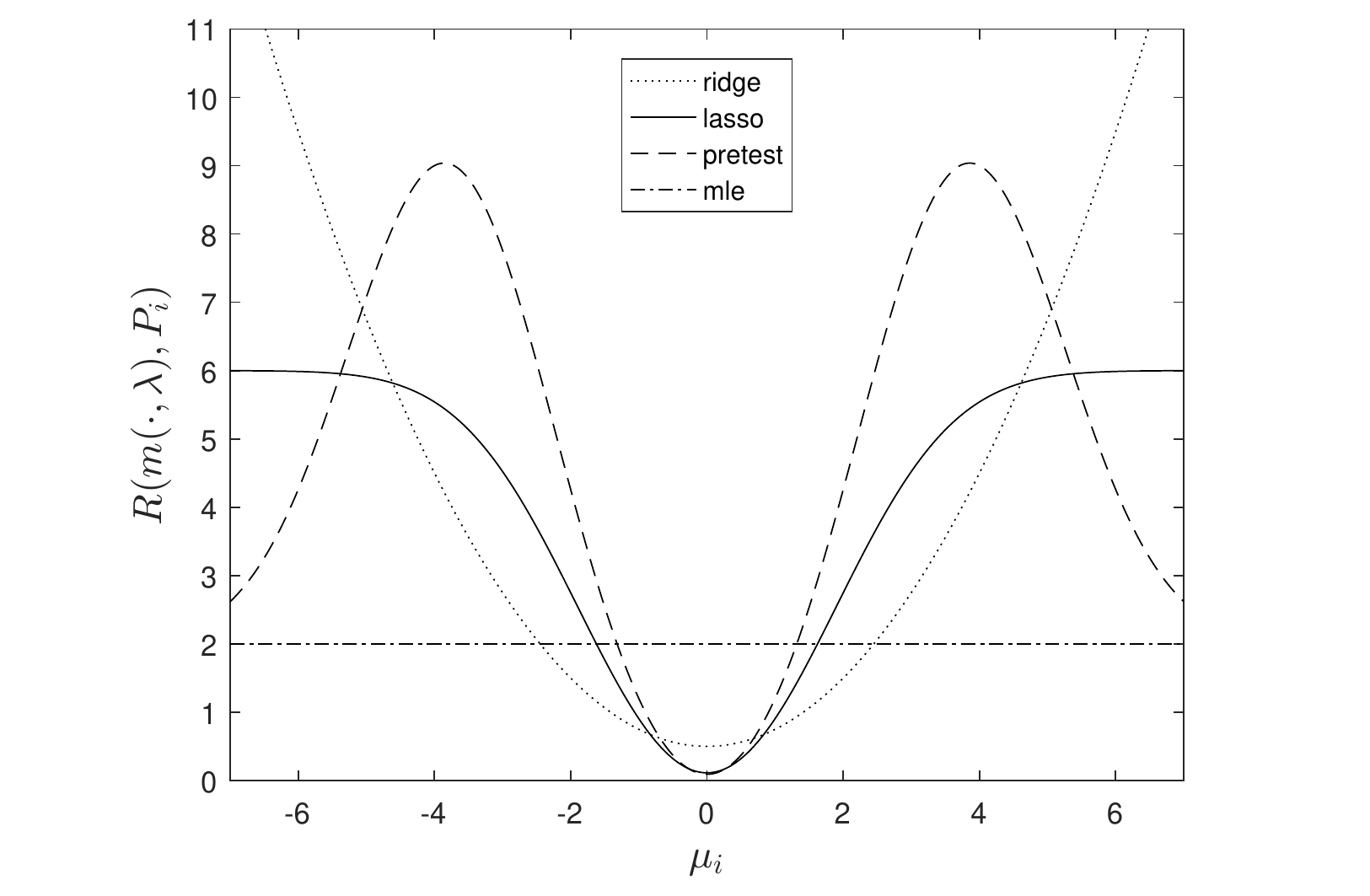}
\end{center}
\floatfoot{\footnotesize This figure displays componentwise risk, $R(m(\cdot,\lambda))$, as a function of $\mu_i$ for componentwise estimators, where $\sigma_i^2=2$. ``mle'' refers to the maximum likelihood (unregularized) estimator, $\widehat \mu_i=X_i$, which has risk equal to $\sigma_i^2=2$. The regularization parameters are $\lambda=1$ for ridge, $\lambda=2$
for lasso, and $\lambda=4$ for pretest, as in Figure \ref{fig:componetwisem}.}
\end{figure}

Figure \ref{fig:comprisk} plots the componentwise risk functions in Lemma \ref{lem:componentwise} as functions of $\mu_i$ (with $\lambda=1$ for ridge, $\lambda=2$ for lasso, and $\lambda=4$ for pretest). It also plots the componentwise risk of the unregularized maximum likelihood estimator, $\widehat\mu_i=X_i$, which is equal to $\sigma_i^2$. As Figure \ref{fig:comprisk} suggests, componentwise risk is large for ridge when $|\mu_i|$ is large. The same is true for lasso, except that risk remains bounded. For pretest, componentwise risk is large when $|\mu_i|$ is close to $\lambda$.

Notice that these functions are plotted for a \textit{fixed} value of the regularization parameter. If $\lambda$ is chosen \textit{optimally} , then the componentwise risks of ridge, lasso, and pretest are no greater than the componentwise risk of the unregularized maximum likelihood estimator $\widehat{\mu}_i=X_i$, which is $\sigma_i^2$. The reason is that ridge, lasso, and pretest nest the unregularized estimator (as the case $\lambda=0$).

\subsection{Spike and normal data generating process}
\label{ssec:spikeandnormal}

If we take the expressions for componentwise risk derived in Lemma \ref{lem:componentwise} and average them over some population distribution of $(\mu_i,\sigma^2_i)$, we obtain the integrated, or empirical Bayes, risk. For parametric families of distributions of $(\mu_i,\sigma^2_i)$, this might be done analytically. We shall do so now, considering a family of distributions that is rich enough to cover common intuitions about data generating processes, but simple enough to allow for analytic expressions. Based on these expressions, we characterize scenarios that favor the relative performance of each of the estimators considered in this article.

We consider a family of distributions for $(\mu_i,\sigma_i)$ such that: (i) $\mu_i$ takes value zero with probability $p$ and is otherwise distributed as a normal with mean value $\mu_0$ and standard deviation $\sigma_0$, and (ii) $\sigma_i^2=\sigma^2$. The following proposition derives the optimal estimating function $\bar m^*_\pi$, as well as integrated risk functions for this family of distributions.

\begin{proposition}[Spike and normal data generating process]
\label{prop:spikenormal}
Assume $\pi$ is such that (i) $\mu_1, \ldots, \mu_n$ are drawn independently from a distribution with probability mass $p$ at zero, and normal with mean $\mu_0$ and variance $\sigma_0^2$ elsewhere, and (ii) conditional on $\mu_i$,  $X_i$ follows a normal distribution with mean $\mu_i$ and variance $\sigma^2$. Then, the optimal shrinkage function is
\[
\bar m^*_\pi(x)=\frac{(1-p)\displaystyle\frac{1}{\sqrt{\sigma_0^2+\sigma^2}}
\phi\left(\displaystyle\frac{x-\mu_0}{\sqrt{\sigma_0^2+\sigma^2}}\right) \displaystyle\frac{\mu_0\sigma^2+x\sigma_0^2}{\sigma_0^2+\sigma^2} }{p\displaystyle\frac{1}{\sigma}\phi\left(\displaystyle\frac{x}{\sigma}\right)+(1-p)\displaystyle\frac{1}{\sqrt{\sigma_0^2+\sigma^2}}
\phi\left(\displaystyle\frac{x-\mu_0}{\sqrt{\sigma_0^2+\sigma^2}}\right)}.
\]
The integrated risk of ridge is
\[
\bar R(m_R(\cdot,\lambda),\pi)=\Bigg(\frac{1}{1+\lambda}\Bigg)^2\sigma^2+(1-p)\Bigg(\frac{\lambda}{1+\lambda}\Bigg)^2(\mu_0^2+\sigma_0^2),
\]
with
\[
\bar\lambda_R^*(\pi)=\frac{\sigma^2}{(1-p)(\mu_0^2+\sigma_0^2)}.
\]
The integrated risk of lasso is given by
\[
\bar R(m_L(\cdot,\lambda),\pi)=p\bar R_{0}(m_{L}(\cdot,\lambda),\pi)+(1-p)\bar R_{1}(m_{L}(\cdot,\lambda),\pi),
\]
where
\[
\bar R_{0}(m_{L}(\cdot,\lambda),\pi)=2\Phi\Big(\displaystyle\frac{-\lambda}{\sigma}\Big)(\sigma^2+\lambda^2)
-2\Big(\displaystyle\frac{\lambda}{\sigma}\Big)\phi\Big(\displaystyle\frac{\lambda}{\sigma}\Big)\sigma^2,
\]
and
\begin{align*}
\bar R_{1}(m_{L}(\cdot,\lambda),\pi)=\Bigg(&1+\Phi\Bigg(\displaystyle\frac{-\lambda-\mu_0}{\sqrt{\sigma_0^2+\sigma^2}}\Bigg)
-\Phi\Bigg(\displaystyle\frac{\lambda-\mu_0}{\sqrt{\sigma_0^2+\sigma^2}}\Bigg)\Bigg)(\sigma^2+\lambda^2)\\
&+ \Bigg(\Phi\Bigg(\displaystyle\frac{\lambda-\mu_0}{\sqrt{\sigma_0^2+\sigma^2}}\Bigg)
-\Phi\Bigg(\displaystyle\frac{-\lambda-\mu_0}{\sqrt{\sigma_0^2+\sigma^2}}\Bigg)\Bigg)(\mu_0^2+\sigma_0^2)\\
&-\frac{1}{\sqrt{\sigma_0^2+\sigma^2}}\phi\Bigg(\displaystyle\frac{\lambda-\mu_0}{\sqrt{\sigma_0^2+\sigma^2}}\Bigg)(\lambda+\mu_0)(\sigma_0^2+\sigma^2)\\
&-\frac{1}{\sqrt{\sigma_0^2+\sigma^2}}\phi\Bigg(\displaystyle\frac{-\lambda-\mu_0}{\sqrt{\sigma_0^2+\sigma^2}}\Bigg)(\lambda-\mu_0)(\sigma_0^2+\sigma^2).
\end{align*}
Finally, the integrated risk of pretest is given by
\[
\bar R(m_{PT}(\cdot,\lambda),\pi)=p\bar R_{0}(m_{PT}(\cdot,\lambda),\pi)+(1-p)\bar R_{1}(m_{PT}(\cdot,\lambda),\pi),
\]
where
\[
\bar R_{0}(m_{PT}(\cdot,\lambda),\pi)=2\Phi\Big(\displaystyle\frac{-\lambda}{\sigma}\Big)\sigma^2
+2\Big(\displaystyle\frac{\lambda}{\sigma}\Big)\phi\Big(\displaystyle\frac{\lambda}{\sigma}\Big)\sigma^2
\]
and
\begin{align*}
\bar R_{1}(m_{PT}(\cdot,\lambda),\pi)=\Bigg(&1+\Phi\Bigg(\displaystyle\frac{-\lambda-\mu_0}{\sqrt{\sigma_0^2+\sigma^2}}\Bigg)
-\Phi\Bigg(\displaystyle\frac{\lambda-\mu_0}{\sqrt{\sigma_0^2+\sigma^2}}\Bigg)\Bigg)\sigma^2\\
&+ \Bigg(\Phi\Bigg(\displaystyle\frac{\lambda-\mu_0}{\sqrt{\sigma_0^2+\sigma^2}}\Bigg)
-\Phi\Bigg(\displaystyle\frac{-\lambda-\mu_0}{\sqrt{\sigma_0^2+\sigma^2}}\Bigg)\Bigg)(\mu_0^2+\sigma_0^2)\\
&-\frac{1}{\sqrt{\sigma_0^2+\sigma^2}}\phi\Bigg(\displaystyle\frac{\lambda-\mu_0}{\sqrt{\sigma_0^2+\sigma^2}}\Bigg)\big(\lambda(\sigma_0^2-\sigma^2)+\mu_0(\sigma_0^2+\sigma^2)\big)\\
&-\frac{1}{\sqrt{\sigma_0^2+\sigma^2}}\phi\Bigg(\displaystyle\frac{-\lambda-\mu_0}{\sqrt{\sigma_0^2+\sigma^2}}\Bigg)\big(\lambda(\sigma_0^2-\sigma^2)-\mu_0(\sigma_0^2+\sigma^2)\big).
\end{align*}
\end{proposition}

Notice that, even under substantial sparsity (that is, if $p$ is large), the optimal shrinkage function, $\bar m^*_\pi$, never shrinks all the way to zero (unless, of course, $\mu_0=\sigma_0=0$ or $p=1$). This could in principle cast some doubts about the appropriateness of thresholding estimators, such as lasso or pretest, which induce sparsity in the estimated parameters. However, as we will see below, despite this stark difference between thresholding estimators and $\bar m^*_\pi$, lasso and, to a certain extent, pretest are able to approximate the integrated risk of $\bar m^*_\pi$ in the spike and normal model when the degree of sparsity in the parameters of interest is substantial.

\begin{figure}\begin{center}
\caption{{Risk for estimators in spike and normal setting}}$ $\\
\label{fig:RiskSpikeNormal}
\footnotesize

\makebox[\linewidth][c]{
\fbox{\includegraphics[width = 0.55\textwidth]{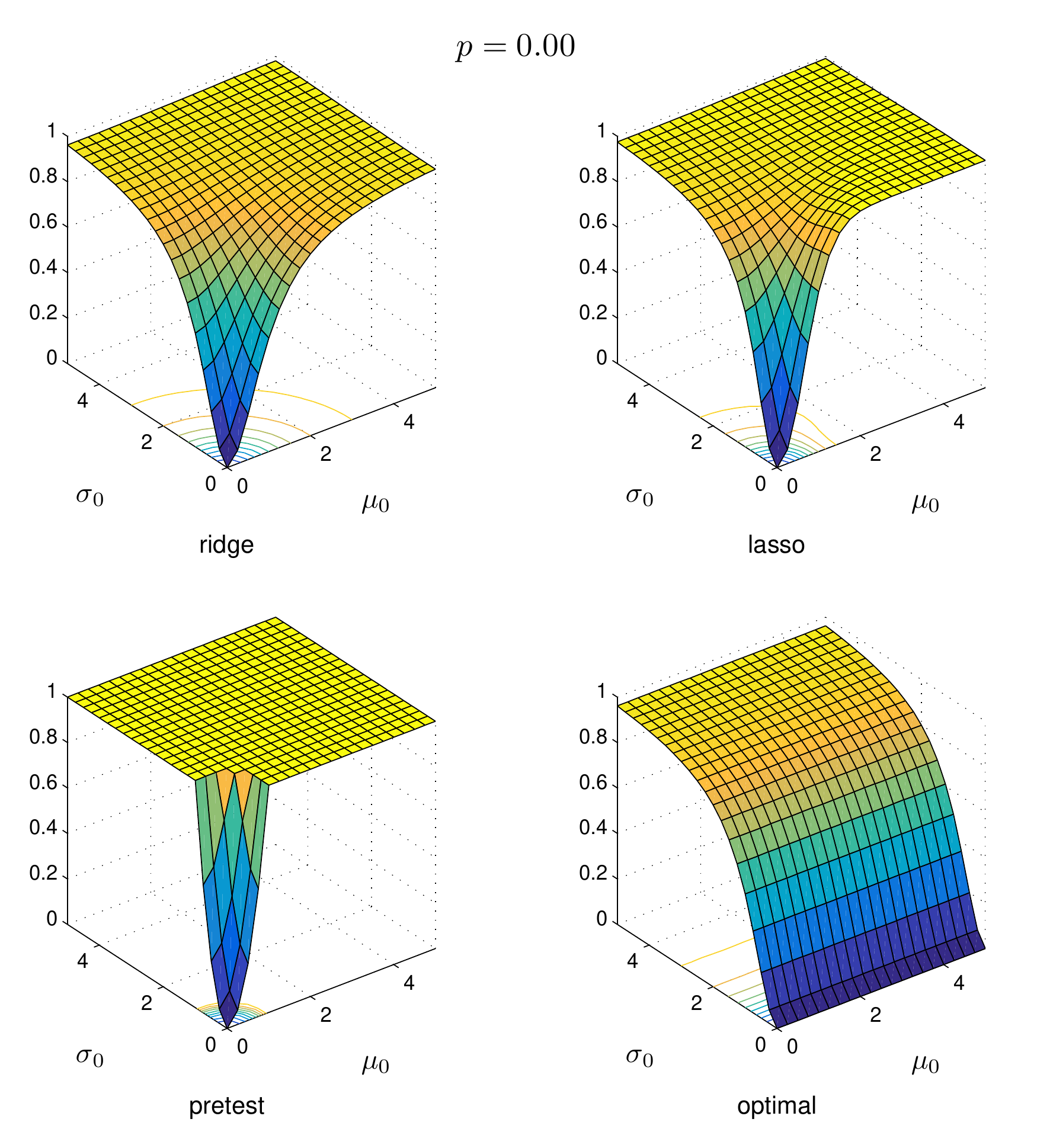}}\quad
\fbox{\includegraphics[width = 0.55\textwidth]{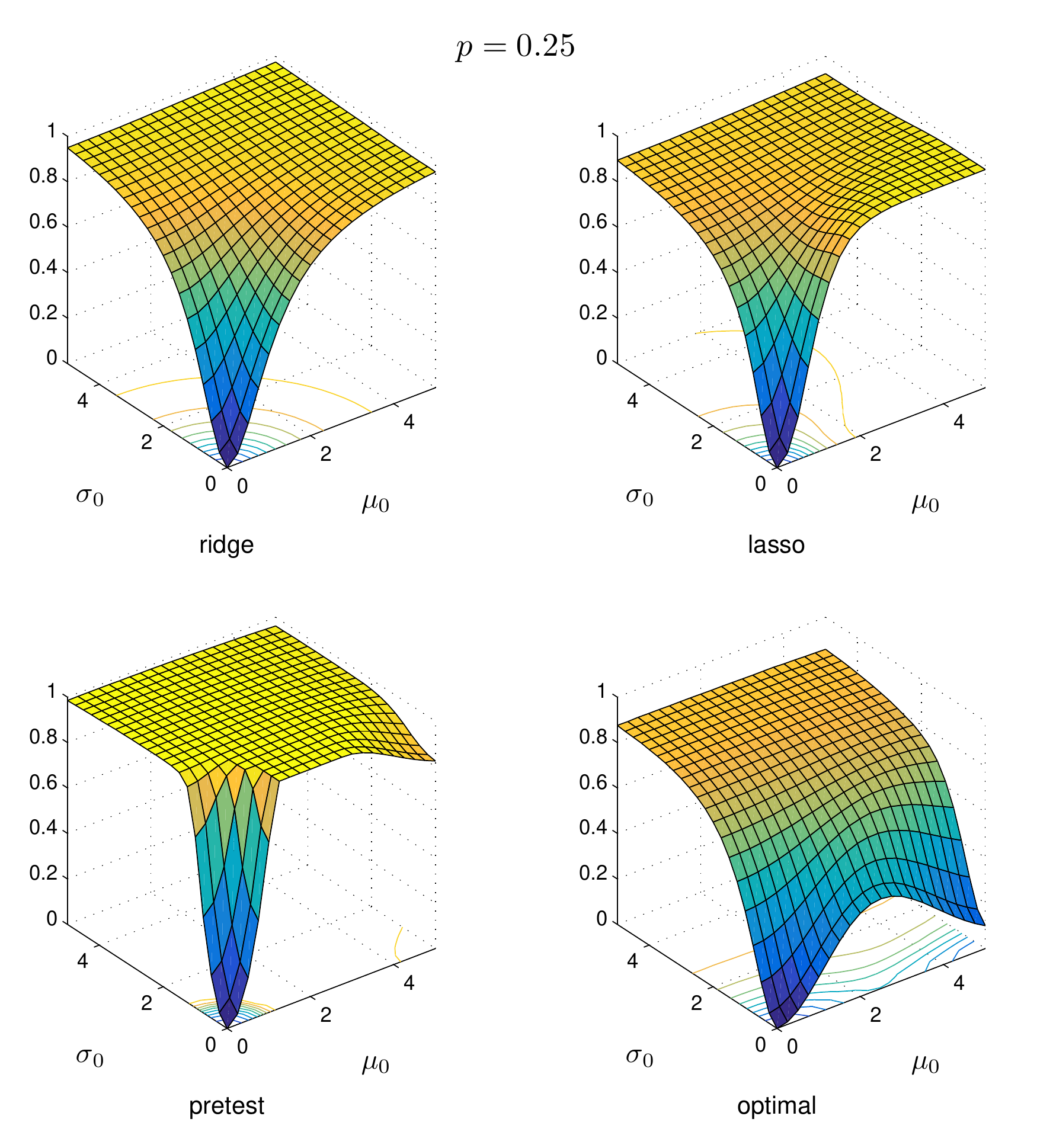}}
}
\smallskip

\makebox[\linewidth][c]{
\fbox{\includegraphics[width = 0.55\textwidth]{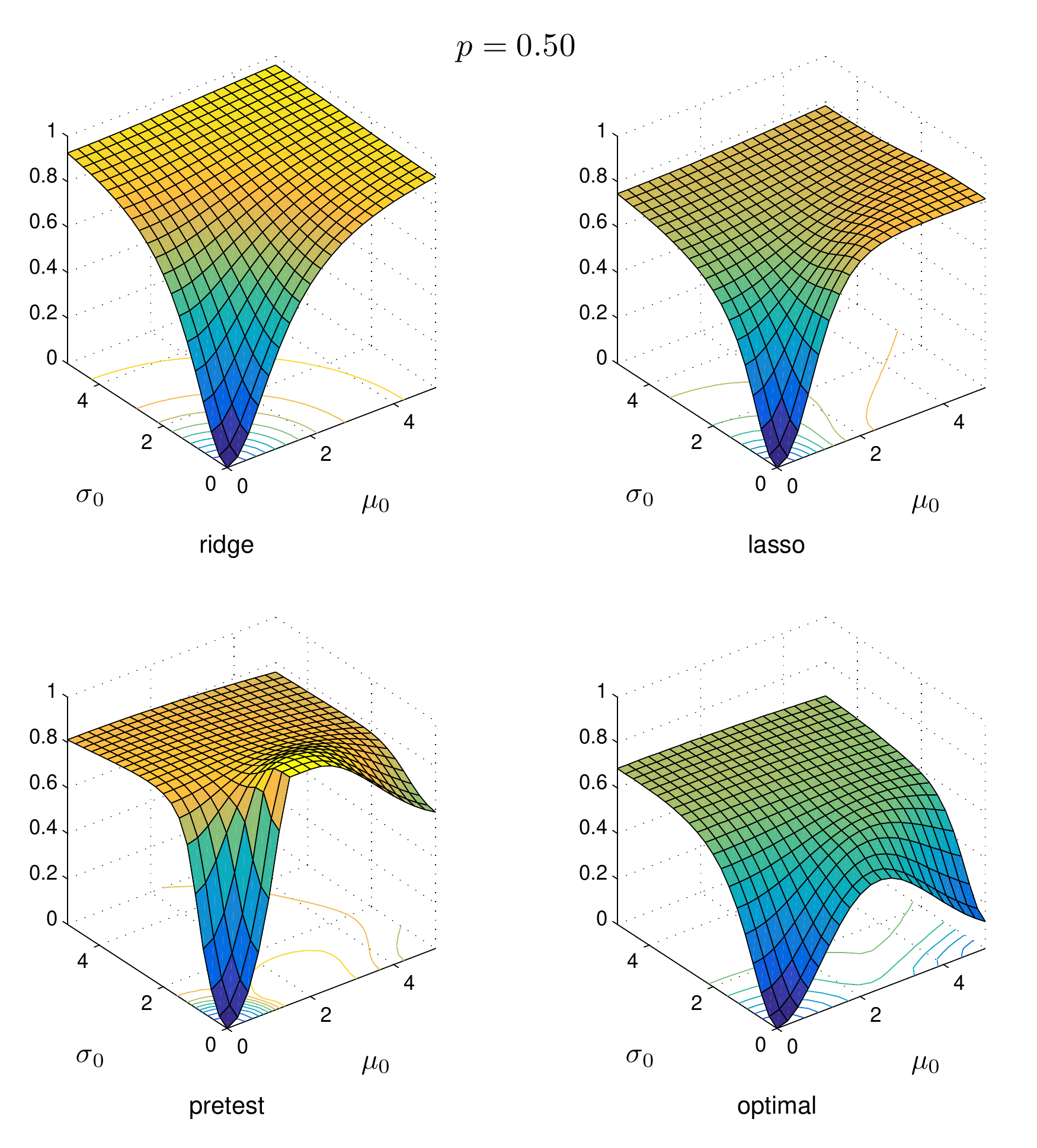}}\quad
\fbox{\includegraphics[width = 0.55\textwidth]{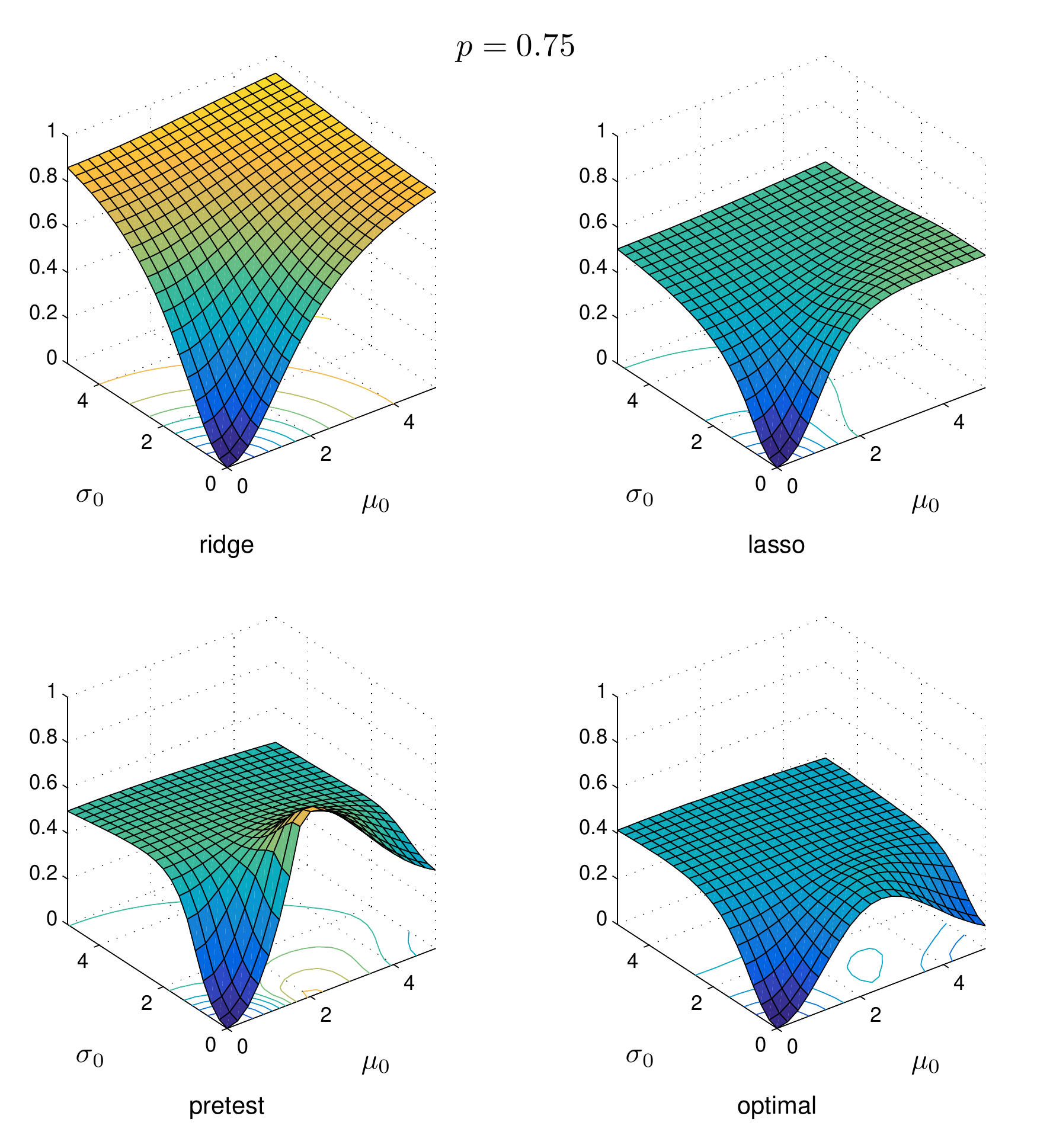}}
}
\end{center}
\end{figure}

\begin{figure}\begin{center}
\caption{{Best estimator in spike and normal setting}}$ $\\
\label{fig:BestSpikeNormal}
\footnotesize
\includegraphics[width = \textwidth]{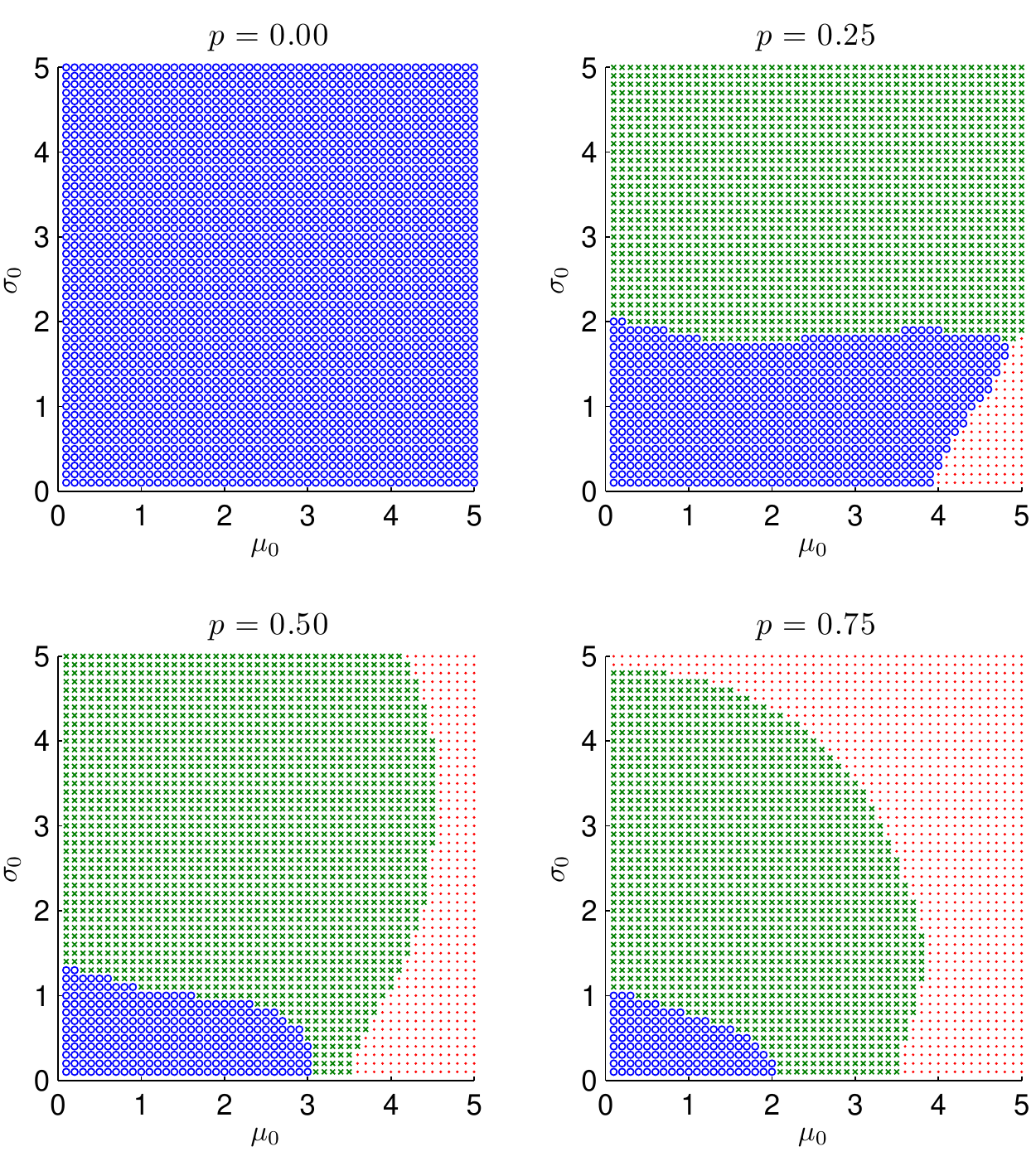}
\end{center}
\floatfoot{\footnotesize This figure compares integrated risk values attained by ridge, lasso, and pretest for different parameter values of the spike and normal specification in Section \ref{ssec:spikeandnormal}. Blue circles are placed at parameters values for which ridge minimizes integrated risk, green crosses at values for which lasso minimizes integrated risk, and red dots are parameters values for which pretest minimizes integrated risk.}
\end{figure}

\paragraph{Visual representations}
While it is difficult to directly interpret the risk formulas in Proposition \ref{prop:spikenormal}, plotting these formulas as functions of the parameters governing the data generating process elucidates some crucial aspects of the risk of the corresponding estimators. Figure \ref{fig:RiskSpikeNormal} does so, plotting the minimal integrated risk function of the different estimators. Each of the four subplots in Figure \ref{fig:RiskSpikeNormal} is based on a fixed value of $p\in\{0,0.25,0.5,0.75\}$, with $\mu_0$ and $\sigma_0^2$ varying along the bottom axes. For each value of the triple $(p,\mu_0,\sigma_0)$, Figure \ref{fig:RiskSpikeNormal} reports minimal integrated risk of each estimator (minimized over $\lambda\in[0,\infty]$). As a benchmark, Figure \ref{fig:RiskSpikeNormal} reports the risk of the optimal shrinkage function, $\bar m^*_\pi$, simulated over 10 million repetitions. Figure \ref{fig:BestSpikeNormal} maps the regions of parameter values over which each of the three estimators, ridge, lasso, or pretest, performs best in terms of integrated risk.

Figures \ref{fig:RiskSpikeNormal} and \ref{fig:BestSpikeNormal} provide some useful insights on the performance of shrinkage estimators. With no true zeros, ridge performs better than lasso or pretest. A clear advantage of ridge in this setting is that, in contrast to lasso or pretest, ridge allows shrinkage without shrinking some observations all the way to zero. As the share of true zeros increases, the relative performance of ridge deteriorates for pairs $(\mu_0,\sigma_0)$ away from the origin. Intuitively, linear shrinkage imposes a disadvantageous trade-off on ridge. Using ridge to heavily shrink towards the origin in order to fit potential true zeros produces large expected errors for observations with $\mu_i$ away from the origin. As a result, ridge performance suffers considerably unless much of the probability mass of the distribution of $\mu_i$ is tightly concentrated around zero. In the absence of true zeros, pretest performs particularly poorly unless the distribution of $\mu_i$ has much of its probability mass tightly concentrated around zero, in which case shrinking all the way to zero produces low risk. However, in the presence of true zeros, pretest performs well when much of the probability mass of the distribution of $\mu_i$ is located in a set that is well-separated from zero, which facilitates the detection of true zeros. Intermediate values of $\mu_0$ coupled with moderate values of $\sigma_0$ produces settings where the conditional distributions $X_i|\mu_i=0$ and $X_i|\mu_i\neq 0$ greatly overlap, inducing substantial risk for pretest estimation. The risk performance of lasso is particularly robust. It out-performs ridge and pretest for values of $(\mu_0,\sigma_0)$ at intermediate distances to the origin, and uniformly controls risk over the parameter space. This robustness of lasso may explain its popularity in empirical practice. Despite the fact that, unlike optimal shrinkage, thresholding estimators impose sparsity, lasso -- and to a certain extent -- pretest are able to approximate the integrated risk of the optimal shrinkage function over much of the parameter space.

All in all, the results in Figures \ref{fig:RiskSpikeNormal} and \ref{fig:BestSpikeNormal} for the spike and normal case support the adoption of ridge in empirical applications where there are no reasons to presume the presence of many true zeros among the parameters of interest. In empirical settings where many true zeros may be expected, Figures \ref{fig:RiskSpikeNormal} and \ref{fig:BestSpikeNormal} show that the choice among estimators in the spike and normal model depends on how well separated the distributions $X_i|\mu_i=0$ and $X_i|\mu_i\neq 0$ are. Pretest is preferred in the well-separated case, while lasso is preferred in the non-separated case.

\section{Data-driven choice of regularization parameters}
\label{sec:datadrivenregularization}

In Section \ref{ssec:spikeandnormal} we adopted a parametric model for the distribution of $\mu_i$ to study the risk properties of regularized estimators under an oracle choice of the regularization parameter, $\bar \lambda^*(\pi)$. In this section, we return to a nonparametric setting and show that it is possible to consistently estimate  $\bar \lambda^*(\pi)$ from the data, $X_1,\ldots , X_n$, under some regularity conditions on $\pi$. We consider estimates $\widehat{\lambda}_n$ of $\bar \lambda^*(\pi)$ based on Stein's unbiased risk estimate and based on cross validation. The resulting estimators $m(X_i,\widehat{\lambda}_n)$ have risk functions which are {\em uniformly} close to those of the infeasible estimators $m(X_i, \bar \lambda^*(\pi))$.

The uniformity part of this statement is important and not obvious.
Absent uniformity, asymptotic approximations might misleadingly suggest good behavior, while in fact the finite sample behavior of proposed estimators might be quite poor for plausible sets of data generating processes.
This uniformity results in this section contrast markedly with other oracle approximations to risk, most notably approximations which assume that the true zeros, that is the components $i$ for which $\mu_i=0$, are known. Asymptotic approximations of this latter form are often invoked when justifying the use of lasso and pretest estimators. Such approximations are in general not uniformly valid, as emphasized by \cite{leeb2005model} and others. 

\subsection{Uniform loss and risk consistency}
\label{ssec:unformrisk}

For the remainder of the paper we adopt the following short-hand notation:
\begin{align*}
L_n(\lambda) &=L_n(\bs X, m(\cdot,\lambda),\bs P)&\textrm{(compound loss)}\\
R_n(\lambda)&=R_n(m(\cdot,\lambda),\bs P)&\textrm{(compound risk)}\\
\bar R_\pi(\lambda)&=\bar R(m(\cdot,\lambda),\pi)&\textrm{(empirical Bayes or integrated risk)}
\end{align*}

We will now consider estimators $\widehat{\lambda}_n$ of $\bar\lambda^*(\pi)$ that are obtained by minimizing some empirical estimate of the risk function $\bar R_\pi$ (possibly up to a constant that depends only on $\pi$). The resulting $\widehat{\lambda}_n$ is then used to obtain regularized estimators of the form $\widehat{\mu}_i = m(X_i, \widehat{\lambda}_n)$. We will show that for large $n$ the compound loss, the compound risk, and the integrated risk functions of the resulting estimators are uniformly close to the corresponding functions of the same estimators evaluated at oracle-optimal values of $\lambda$. As $n\rightarrow \infty$, the differences between $L_n$, $R_n$, and $\bar R_\pi$ vanish, so compound loss optimality, compound risk optimality, and integrated risk optimality become equivalent.

The following theorem establishes our key result for this section. Let $\mathcal Q$ be a set of probability distributions for $(X_i,\mu_i)$. Theorem \ref{theorem:loss} provides sufficient conditions for uniform loss consistency over $\pi\in\mathcal Q$, namely that (i) the supremum of the difference between the loss, $L_n(\lambda)$, and the empirical Bayes risk, $\bar R_\pi(\lambda)$, vanishes in probability uniformly over $\pi\in\mathcal Q$ and (ii) that $\widehat{\lambda}_n$ is chosen to minimize a uniformly consistent estimator, $r_n(\lambda)$, of the risk function, $\bar R_\pi(\lambda)$ (possibly up to a constant $\bar v_\pi$). Under these conditions, the difference between loss $L_n(\widehat\lambda_n)$ and the infeasible minimal loss $\inf_{\lambda\in [0,\infty]} L_n (\lambda)$ vanishes in probability uniformly over $\pi\in\mathcal Q$.

\begin{theorem}[Uniform loss consistency]
\label{theorem:loss}
Assume
\begin{equation}
\label{equation:residual}
\sup_{\pi\in\mathcal Q} P_\pi\left(\sup_{\lambda\in [0,\infty]} \Big|L_n(\lambda)-\bar R_\pi(\lambda)\Big|>\epsilon\right)\rightarrow 0,\quad \forall \epsilon >0.
\end{equation}
Assume also that there are functions, $\bar r_\pi(\lambda)$, $\bar v_\pi$, and $r_n(\lambda)$ (of $(\pi,\lambda)$, $\pi$, and $(\{X_i\}_{i=1}^n,\lambda)$, respectively) such that $\bar R_\pi(\lambda)=\bar r_\pi(\lambda)+\bar v_\pi$, and
\begin{equation}
\label{equation:ullnforr}
\sup_{\pi\in\mathcal Q} P_\pi\left(\sup_{\lambda\in [0,\infty]} \big|r_n(\lambda)-\bar r_\pi(\lambda)\big|>\epsilon\right)\rightarrow 0,\quad \forall \epsilon >0.
\end{equation}
Then,
\begin{equation*}
\label{equation:uniflosscons}
\sup_{\pi\in\mathcal Q} P_\pi\left(\left|L_n(\widehat\lambda_n)-\inf_{\lambda\in [0,\infty]} L_n (\lambda)\right|>\epsilon\right)\rightarrow 0,\quad \forall \epsilon >0,
\end{equation*}
where $\widehat\lambda_n = \argmin_{\!\!\lambda\in [0,\infty]} r_n(\lambda)$.
\end{theorem}

The sufficient conditions given by this theorem, as stated in equations \eqref{equation:residual} and \eqref{equation:ullnforr}, are rather high-level. We shall now give more primitive conditions for these requirements to hold. In Sections \ref{section:SURE} and \ref{ssec:CV} below, we propose suitable choices of $r_n(\lambda)$ based on Stein's unbiased risk estimator (SURE) and cross-validation (CV), and show that equation (\ref{equation:ullnforr}) holds for these choices of $r_n(\lambda)$.

The following Theorem \ref{theorem:l2conv} provides a set of conditions under which equation \eqref{equation:residual} holds, so the difference between compound loss and integrated risk vanishes uniformly. Aside from a bounded moment assumption, the conditions in Theorem \ref{theorem:l2conv} impose some restrictions on the estimating functions, $m(x,\lambda)$. Lemma \ref{lemma:l2rlp} below shows that those conditions hold, in particular, for ridge, lasso, and pretest estimators.

\begin{theorem}[Uniform $L^2$-convergence]
\label{theorem:l2conv}
Suppose that\vspace*{-0.05in}
\begin{enumerate}
\item $m(x,\lambda)$ is monotonic in $\lambda$ for all $x$ in $\mathbb R$,\label{item:monotonic}
\item $m(x,0) = x$ and $\lim_{\lambda\rightarrow \infty} m(x,\lambda) = 0$ for all $x$ in $\mathbb R$,\label{item:limits}
\item $\sup_{\pi\in\mathcal Q} E_\pi[X^4] <\infty$.\label{item:moments}
\item For any $\epsilon>0$ there exists a set of regularization parameters $0=\lambda_0 < \ldots < \lambda_k=\infty$, which may depend on $\epsilon$, such that \label{item:regparam}
\[E_\pi[(|X - \mu| +  |\mu|) |m(X,\lambda_j) - m(X,\lambda_{j-1})|] \leq \epsilon\]
for all $j=1,\ldots, k$ and all $\pi \in \mathcal Q$.
\end{enumerate}
Then,
\begin{equation}
\label{equation:l2conv}
\sup_{\pi\in\mathcal Q} E_\pi\left[\sup_{\lambda\in [0,\infty]} \Big(L_n(\lambda)-\bar R_\pi(\lambda)\Big)^2\right]\rightarrow 0.
\end{equation}
\end{theorem}

Notice that finiteness of $\sup_{\pi\in\mathcal Q} E_\pi[X^4]$ is equivalent to finiteness of $\sup_{\pi\in\mathcal Q} E_\pi[\mu^4]$ and $\sup_{\pi\in\mathcal Q} E_\pi[(X-\mu)^4]$ via Jensen's and Minkowski's inequalities.

\begin{lemma}
\label{lemma:l2rlp}
If $\sup_{\pi\in\mathcal Q} E_\pi[X^4] <\infty$, then equation (\ref{equation:l2conv}) holds for ridge and lasso. If, in addition, $X$ is continuously distributed with a bounded density, then equation (\ref{equation:l2conv}) holds for pretest.
\end{lemma}

Theorem \ref{theorem:loss} provides sufficient conditions for uniform \emph{loss} consistency. The following corollary shows that under the same conditions we obtain uniform \emph{risk} consistency, that is, the integrated risk of the estimator based on the data-driven choice $\widehat{\lambda}_n$ becomes uniformly close to the risk of the oracle-optimal $\bar \lambda^*(\pi)$. For the statement of this corollary, recall that $\bar{R}(m(., \widehat{\lambda}_n), \pi)$ is the integrated risk of the estimator  $m(., \widehat{\lambda}_n)$ using the stochastic (data-dependent) $\widehat{\lambda}_n$.

\begin{corollary}[Uniform risk consistency]
\label{corollary:risk}
Under the assumptions of Theorem \ref{theorem:l2conv},
\begin{equation}
\label{equation:unifRpicons}
\sup_{\pi\in\mathcal Q} \left|\bar{R}(m(., \widehat{\lambda}_n), \pi)-\inf_{\lambda\in [0,\infty]} \bar R_\pi (\lambda)\right|\rightarrow 0.
\end{equation}
\end{corollary}

In this section, we have shown that approximations to the risk function of machine learning estimators  based on oracle-knowledge of $\lambda$ are uniformly valid over $\pi\in\mathcal Q$ under mild assumptions. It is worth pointing out that such uniformity is not a trivial result. This is made clear by comparison to an alternative approximation, sometimes invoked to motivate the adoption of machine learning estimators, based on oracle-knowledge of true zeros among $\mu_1, \ldots, \mu_n$ (see, e.g., \citealt{fan2001oracle}).
As shown in Appendix \ref{asection:oracle}, assuming oracle knowledge of zeros does not yield a uniformly valid approximation.

\subsection{Stein's unbiased risk estimate} \label{section:SURE}

Theorem \ref{theorem:loss} provides sufficient conditions for uniform loss consistency using a general estimator $r_n$ of risk. We shall now establish that our conditions apply to a particular estimator of $r_n$, known as Stein's unbiased risk estimate (SURE), which was first proposed by \cite{stein1981estimation}. SURE leverages the assumption of normality to obtain an elegant expression of risk as an expected sum of squared residuals plus a penalization term.

SURE as originally proposed requires that $m$ be piecewise differentiable as a function of $x$, which excludes discontinuous estimators such as the pretest estimator $m_{PT}(x,\lambda)$. We provide a generalization in Lemma \ref{lem:SURE} that allows for discontinuities. This lemma is stated in terms of integrated risk; with the appropriate modifications, the same result holds verbatim for compound risk.

\begin{lemma}[SURE for piecewise differentiable estimators]
\label{lem:SURE}
Suppose that $\mu\sim \vartheta$ and
\[X | \mu \sim N(\mu,1).\]
Let $f_\pi = \vartheta \ast \phi$ be the marginal density of $X$, where $\phi$ is the standard normal density.
Consider an estimator $m(X)$ of $\mu$, and suppose that $m(x)$ is differentiable everywhere in $\mathbb{R} \backslash \{x_1, \ldots, x_J\}$, but might be discontinuous at $\{x_1, \ldots, x_J\}$. Let $\nabla m$ be the derivative of $m$ (defined arbitrarily at  $\{x_1, \ldots, x_J\}$), and let $\Delta m_j = \lim_{x\downarrow x_j} m(x) - \lim_{x\uparrow x_j} m(x)$ for $j\in \{1,\ldots,J\}$. Assume that $E_\pi[(m(X)-X)^2]<\infty$, $E_\pi[\nabla m(X)]<\infty$, and $(m(x)-x)\phi(x-\mu)\rightarrow 0$ as $|x|\rightarrow \infty$ $\vartheta$-a.s. Then,
\[
\bar R(m(.), \pi)=  E_\pi[(m(X) - X)^2 ]
+ 2 \left (E_\pi[\nabla m(X)] + \sum_{j=1}^J \Delta m_j f_\pi(x_j) \right )-1.
\]
\end{lemma}

The result of this lemma yields an objective function for the choice of $\lambda$ of the general form we considered in Section \ref{ssec:unformrisk}, with $\bar v_\pi=-1$ and
\begin{equation}
\label{equation:rpisure}
\bar r_\pi (\lambda) = E_\pi[(m(X,\lambda) - X)^2] + 2 \left (E_\pi[\nabla_{\!x} m(X,\lambda)] + \sum_{j=1}^J \Delta m_j(\lambda) f_\pi(x_j) \right),
\end{equation}
where $\nabla_{\!x} m(x,\lambda)$ is the derivative of $m(x,\lambda)$ with respect to its first argument, and $\{x_1,\ldots, x_J\}$ may depend on $\lambda$. The expression in equation \eqref{equation:rpisure} can be estimated using its sample analog,
\begin{equation}
\label{equation:surecrit}
r_n(\lambda) = \frac{1}{n}\sum_{i=1}^n (m(X_i,\lambda) - X_i)^2 + 2 \left (\frac{1}{n}\sum_{i=1}^n\nabla_{\!x} m(X_i,\lambda) + \sum_{j=1}^J \Delta m_j(\lambda) \widehat{f}(x_j) \right ),
\end{equation}
where $\widehat f(x)$ is an estimator of $f_\pi(x)$. This expression can be thought of as a penalized least squares objective function. The following are explicit expressions for the penalty for the cases of ridge, lasso, and pretest.\bigskip

\begin{tabular}{lll}
ridge: && $\displaystyle\frac{2}{1+\lambda}$\\[3ex]
lasso: && $\displaystyle\frac{2}{n} \sum_{i=1}^n1(|X_i| > \lambda)$\\[3ex]
prestest: && $\displaystyle\frac{2}{n} \sum_{i=1}^n1(|X_i| > \lambda)+ 2 \lambda (\widehat{f}(-\lambda) + \widehat{f}(\lambda))$
\end{tabular}

The lasso penalty was previously derived in \cite{donoho1995sure}. Our results allow to apply SURE estimation of risk to any machine learning estimator, as long as the conditions of Lemma \ref{lem:SURE} are satisfied.

To apply the uniform risk consistency in Theorem \ref{theorem:loss}, we need to show that equation \eqref{equation:ullnforr} holds. That is, we have to show that $r_n(\lambda)$ is uniformly consistent as an estimator of $\bar r_\pi (\lambda)$. The following lemma provides the desired result.
\begin{lemma}
\label{lemma:consr}
Assume the conditions of Theorem \ref{theorem:l2conv}. Then, equation (\ref{equation:ullnforr}) holds for $m(\cdot,\lambda)$ equal to $m_R(\cdot,\lambda)$, $m_L(\cdot,\lambda)$. If, in addition,
\[
\sup_{\pi\in\mathcal Q} P_\pi\left(\sup_{x\in\mathbb R} \left||x|\widehat f(x)-|x|f_\pi(x)\right|>\epsilon\right)\rightarrow 0\quad
\forall \epsilon>0,
\]
then equation (\ref{equation:ullnforr}) holds for $m(\cdot,\lambda)$ equal to $m_{PT}(\cdot,\lambda)$.
\end{lemma}

\paragraph{Identification of $\bar m^*_\pi$}
Under the conditions of Lemma \ref{lem:SURE} the optimal regularization parameter $\bar \lambda^*(\pi)$ is identified. In fact, under the same conditions, the stronger result holds that $\bar m^*_\pi$ as defined in Section \ref{ssec:generalriskfunction} is identified as well (see, e.g., \citeauthor{brown1971admissible}, \citeyear{brown1971admissible}; \citeauthor{efron2011tweedie},
\citeyear{efron2011tweedie}). The next lemma states the identification result for $\bar m^*_\pi$.
\begin{lemma}
\label{lemma:optshrink}
Under the conditions of Lemma \ref{lem:SURE}, the optimal shrinkage function is given by
\begin{equation}
\label{equation:optshrink}
\bar m^*_\pi(x)=x+\nabla \log (f_\pi(x)).
\end{equation}
\end{lemma}
Several nonparametric empirical Bayes estimators (NPEB) that target $\bar m^*_\pi(x)$ have been proposed (see \citeauthor{brown2009nonparametric}, \citeyear{brown2009nonparametric}; \citeauthor{jiang2009eb}, \citeyear{jiang2009eb}, \citeauthor{efron2011tweedie},
\citeyear{efron2011tweedie}, and \citeauthor{koenker2014convex}, \citeyear{koenker2014convex}). 
In particular,  \cite{jiang2009eb} derive asymptotic optimality results for nonparametric estimation of $\bar m^*_\pi$ and provide an estimator based on the EM-algorithm. 
The estimator proposed in \cite{koenker2014convex}, which is based on convex optimization techniques, is particularly attractive, both in terms of computational properties and because it sidesteps the selection of a smoothing parameters (cf., e.g., \citeauthor{brown2009nonparametric}, \citeyear{brown2009nonparametric}). Both estimators, in \cite{jiang2009eb} and \cite{koenker2014convex}, use a discrete distribution over a finite number of values to approximate the true distribution of $\mu$. In sections \ref{section:simulations} and \ref{sec:applications}, we will use the Koenker-Mizera estimator to visually compare the shape of this estimated $\bar m^*_\pi(x)$ to the shape of ridge, lasso and pretest estimating functions and to assess the performance of ridge, lasso and pretest relative to the performance of a nonparametric estimator of  $\bar m^*_\pi$.

\subsection{Cross-validation}
\label{ssec:CV}

A popular alternative to SURE is cross-validation, which chooses tuning parameters to optimize out-of-sample prediction.
In this section, we investigate data-driven choices of the regularization parameter in a panel data setting, where multiple observations are available for each value of $\mu$ in the sample.

For $i=1,\ldots, n$, consider i.i.d. draws, $(x_{1i},\ldots, x_{ki},\mu_i,\sigma_i)$, of a random variable $(x_1,\ldots, \allowbreak x_{k},\mu,\sigma)$ with distribution $\pi\in\mathcal Q$ . Assume that the components of $(x_{1},\ldots, x_{k})$ are i.i.d. conditional on $(\mu, \sigma^2)$ and that for each $j=1,\ldots, k$,
\begin{align*}
E[x_j|\mu, \sigma] &= \mu,\\ \
\mbox{var}(x_j|\mu, \sigma) &= \sigma^2.
\end{align*}
Let
\[
X_k=\frac{1}{k} \sum_{j=1}^k x_{j}\quad\mbox{ and }\quad X_{ki}=\frac{1}{k} \sum_{j=1}^k x_{ji}.
\]
For concreteness and to simplify notation, we will consider an estimator based on the first $k-1$ observations for each group $i=1,\ldots, n$,
\[
\widehat{\mu}_{k-1i}=m(X_{k-1i}, \lambda),
\]
and will use observations $x_{ki}$, for $i=1,\ldots n$, as a hold-out sample to choose $\lambda$. Similar results hold for alternative sample partitioning choices.
The loss function and empirical Bayes risk function of this estimator are given by
\begin{align*}
L_{n,k}(\lambda) &= \frac{1}{n}\sum_{i=1}^n (m(X_{k-1i}, \lambda) - \mu_i)^2\\ \intertext{and}
\bar R_{\pi,k}(\lambda)&= E_\pi[(m(X_{k-1}, \lambda) - \mu)^2 ].
\end{align*}
Consider the following cross-validation estimator
\[
r_{n,k}(\lambda) = \frac{1}{n} \sum_{i=1}^n \left (m(X_{k-1i},\lambda)- x_{ki}\right )^2.
\]
\begin{lemma}
\label{lemma:cv}
Assume Conditions \ref{item:monotonic} and \ref{item:limits} of Theorem \ref{theorem:l2conv} and $E_\pi[x_{j}^2]<\infty$, for $j=1,\ldots k$. Then,
\[
E_\pi[r_{n,k}(\lambda)]= \bar  R_{\pi,k}(\lambda)+E_\pi[\sigma^2].
\]
\end{lemma}

That is, the cross validation yields an (up to a constant) unbiased estimator for the risk of the estimating function $m(X_{k-1},\lambda)$. The following theorem shows that this result can be strengthened to a uniform consistency result.
\begin{theorem}
\label{theorem:cvunif}
Assume conditions \ref{item:monotonic} and \ref{item:limits} of Theorem \ref{theorem:l2conv} and $\sup_{\pi} E_\pi[x_{j}^4]<\infty$, for $j=1,\ldots k$. Let $\bar v_\pi = -E_\pi[\sigma^2]$,
\begin{align*}
\bar r_{\pi,k}(\lambda) &= E_\pi[r_{n,k}(\lambda)],\\
&=\bar R_{\pi,k}(\lambda) -\bar v_\pi,
\end{align*}
and $\widehat\lambda_n= \argmin_{\lambda\in [0,\infty]} r_{n,k}(\lambda)$. Then, for ridge, lasso, and pretest,
\[
\sup_{\pi\in\mathcal Q} E_\pi\left[\sup_{\lambda\in [0,\infty]} \Big(r_{n,k}(\lambda)-\bar r_{\pi,k}(\lambda)   \Big)^2\right]\rightarrow 0,
\]
and
\[
\sup_{\pi\in\mathcal Q} P_\pi\left(\left|L_{n,k}\left (\widehat\lambda_n\right) -  \inf_{\lambda\in [0,\infty]} L_{n,k} \left(\lambda\right)\right|>\epsilon\right)\rightarrow 0,\quad \forall \epsilon >0.
\]
\end{theorem}\bigskip

Cross-validation has advantages as well as disadvantages relative to SURE. On the positive side, cross-validation does not rely on normal errors, while SURE does. Normality is less of an issue if $k$ is large, so $X_{ki}$ is approximately normal. On the negative side, however, cross-validation requires holding out part of the data from the second step estimation of $\bs\mu$, once the value of the regularization parameter has been chosen in a first step. This affects the essence of the cross-validation efficiency results, which apply to estimators of the form $m(X_{k-1i},\lambda)$, rather than to feasible estimators that use the entire sample in the second step, $m(X_{ki},\lambda)$. Finally, cross-validation imposes greater data availability requirements, as it relies on  availability of data on repeated realizations, $x_{1i},\ldots, x_{ki}$, of a random variable centered at $\mu_i$, for each sample unit $i=1,\ldots, n$. This may hinder the practical applicability of cross-validation selection of regularization parameters in the context considered in this article.

\section{Discussion and Extensions}
\label{sec:discussion}

\subsection{Mixed estimators and estimators of the optimal shrinkage function}
\label{sec:mixed}

We have discussed criteria such as SURE and CV as means to select the regularization parameter, $\lambda$. In principle, these same criteria might also be used to choose among alternative estimators, such as ridge, lasso, and pretest, in specific empirical settings. Our uniform risk consistency results imply that such a mixed-estimator approach dominates each of the estimators which are being mixed, for $n$ large enough. Going even further, one might aim to estimate the optimal shrinkage function, $\bar m_\pi^*$, using the result of Lemma \ref{lemma:optshrink}, as in \cite{jiang2009eb}, \cite{koenker2014convex}) and others. 
Under suitable consistency conditions, this approach will dominate all other componentwise estimators for large enough $n$ (\citeauthor{jiang2009eb}, \citeyear{jiang2009eb}). In practice, these results should be applied with some caution, as they are based on neglecting the variability in the choice of estimation procedure or in the estimation of $\bar m_\pi^*$. For small and moderate values of $n$, procedures with fewer degrees of freedom may perform better in practice. We return to this issue in section \ref{section:simulations}, where we compare the finite sample risk of the machine learning estimators considered in this article (ridge, lasso and pretest) to the finite sample risk of the NPEB estimator of \cite{koenker2014convex}.

\subsection{Heteroskedasticity}

While for simplicity many of our results are stated for the homoskedastic case, where $\mbox{var}(X_i)= \sigma$ for all $i$, they easily generalize to heteroskedasticity.

The general characterization of compound risk in Theorem \ref{theo:mstar} does not use homoskedasticity, nor does the derivation of componentwise risk in Lemma \ref{lem:componentwise}. The analytical derivations of empirical Bayes risk for the spike and normal data generating process in Proposition \ref{prop:spikenormal}, and the corresponding comparisons of risk in Figures \ref{fig:RiskSpikeNormal} and \ref{fig:BestSpikeNormal} do rely on homoskedasticity. Similar formulas to those of Proposition \ref{prop:spikenormal} might be derived for other data generating processes with heteroskedasticity, but the rankings of estimators might change.

As for our proofs of uniform risk consistency, our general results (Theorem \ref{theorem:loss} and \ref{theorem:l2conv}) do not require homoskedasticity, nor does the validity or consistency of crossvalidation, cf. Theorem \ref{theorem:cvunif}. SURE, in the form we introduced in Lemma \ref{lem:SURE}, does require homoskedasticity. However, the definition of SURE, and the corresponding consistency results, can be extended to the heteroskedastic case (see \citeauthor{xie2012sure}, \citeyear{xie2012sure}).

\subsection{Comparison with \cite{leeb2006risk}}
\label{sec:comparison}

Our results on the uniform consistency of estimators of risk such as SURE or CV appear to stand in contradiction to those of \cite{leeb2006risk}. They consider the same setting as we do -- estimation of normal means -- and the same types of estimators, including ridge, lasso, and pretest. In this setting, \cite{leeb2006risk} show that no uniformly consistent estimator of risk exists for such estimators.

The apparent contradiction between our results and the results in \cite{leeb2006risk} is explained by the different nature of the asymptotic sequence adopted in this article to study the properties of machine learning estimators, relative to the asymptotic sequence adopted in \cite{leeb2006risk} for the same purpose.
In this article, we consider the problem of estimating a large number of parameters, such as location effects for many locations or group-level treatment effects for many groups. This motivates the adoption of an asymptotic sequence along which the number of estimated parameters increases as $n\rightarrow\infty$.
In contrast, \cite{leeb2006risk} study the risk properties of regularized estimators embedded in a sequence along which the number of estimated parameters stays fixed as $n\rightarrow\infty$ and the estimation variance is of order $1/n$.
We expect our approximation to work well when the dimension of the estimated parameter is large; the approximation of \cite{leeb2006risk} is likely to be more appropriate when the dimension of the estimated parameter is small while sample size is large.

In the simplest version of the setting in \cite{leeb2006risk} we observe a $(k\times 1)$ vector $\bs X_n$ with distribution $\bs X_n\sim N(\bs \mu_n, \bs I_{k}/n)$,
where $\bs I_{k}$ is the identity matrix of dimension $k$.
Let $X_{ni}$ and $\mu_{ni}$ be the $i$-components of $\bs X_n$ and $\bs\mu_n$, respectively. Consider the componentwise estimator $m_n(X_{ni})$ of $\mu_{ni}$.
\cite{leeb2006risk} study consistent estimation of the normalized risk
\[
\bar R_n^{LP} = n E\|\bs m_n(\bs X_n)-\bs\mu_n\|^2,
\]
where $\bs m_n(\bs X_n)$ is a $(k\times 1)$ vector with $i$-th element equal to $m_n(X_{ni})$.

Adopting the re-parametrization, $\bs Y_n =\sqrt n \bs X_n$ and $\bs h_n = \sqrt n \bs\mu_n$, we obtain
$\bs Y_n-\bs h_n \sim N(\bs 0, \bs I_k)$. Notice that, for the maximum likelihood estimator, $\bs m_n(\bs X_n)-\bs\mu_n=(\bs Y_n-\bs h_n)/\sqrt n$ and $\bar R_n^{LP} = E|\|m(\bs Y_n)-\bs h_n\|^2=k$, so the risk of the maximum likelihood estimator does not depend on the sequence $\bs h_n$ and, therefore, can be consistently estimated. This is not the case for shrinkage estimators, however. Choosing $\bs h_n=\bs h$ for some fixed $\bs h$, the problem becomes invariant in $n$,
\[
\bs Y_n\sim N(\bs h, \bs I_k).
\]
In this setting, it is easy to show that the risk of machine learning estimators, such as ridge, lasso, and pretest depends on $\bs h$, and therefore it cannot be estimated consistently. For instance, consider the lasso estimator, $m_n(x)=m_{L}(x,\lambda_n)$, where $\sqrt n \lambda_n \rightarrow c$ with $0<c<\infty$, as in \cite{leeb2006risk}. Then, Lemma \ref{lem:componentwise} implies that $\bar R_n^{LP}$ is constant in $n$ and dependent on $\bs h$. As a result, $\bar R_n^{LP}$ cannot be estimated consistently.\footnote{This result holds more generally outside the normal error model. Let $\bs m_L(\bs X_n,\lambda)$ be the $(n\times 1)$ vector with $i$-th element equal to $m_L(X_{i},\lambda)$. Consider the sequence of regularization parameters $\lambda_n=c/\sqrt n$, then $m_{L}(x,\lambda_n)=m_{L}(\sqrt n x,c)/\sqrt n$. This implies $\bar R_n^{LP}=E|\|\bs m_L(\bs Y_n,c)-\bs h\|^2$, which is invariant in $n$.}

Contrast the setting in \cite{leeb2006risk} to the one adopted in this article, where we consider a high dimensional setting, such that $\bs X$ and $\bs \mu$ have dimension equal to $n$. The pairs $(X_i,\mu_i)$ follow a distribution $\pi$ which may vary with $n$. As $n$ increases, $\pi$ becomes identified and so does the average risk, $E_\pi[(m_n(X_i)-\mu_i)^2]$, of any componentwise estimator, $m_n(\cdot)$.

Whether the asymptotic approximation in \cite{leeb2006risk} or ours provides a better description of the performance of SURE, CV, or other estimators of risk in actual applications depends on the dimension of $\bs\mu$. If this dimension is large, as typical in the applications we consider in this article, we expect our uniform consistency result to apply: a ``blessing of dimesionality''. As demonstrated by Leeb and  P{\"o}tscher, however, precise estimation of a fixed number of parameters does not ensure uniformly consistent estimation of risk.

\section{Simulations}
\label{section:simulations}

\paragraph{Designs}
To gauge the relative performance of the estimators considered in this article, we next report the results of a set of simulations that employ the spike and normal data generating process of Section \ref{ssec:spikeandnormal}. As in Proposition \ref{prop:spikenormal}, we consider distributions $\pi$ of $(X,\mu)$ such that $\mu$ is degenerate at zero with probability $p$ and normal with mean $\mu_0$ and variance $\sigma_0^2$ with probability $(1-p)$. 
We consider all combinations of parameter values $p=0.00, 0.25, 0.50, 0.75,\allowbreak 0.95$, $\mu_0=0, 2, 4$, $\sigma_0=2, 4, 6$, and sample sizes $n=50, 200, 1000$.  

Given a set of values $\mu_1,\ldots, \mu_n$, the values for $X_1, \ldots, X_n$ are generated as follows. To evaluate the performance of estimators based on SURE selectors and of the NPEB estimator of \cite{koenker2014convex}, we generate the data as
\begin{equation}
\label{equation:Xi}
X_i = \mu_i + U_i,
\end{equation}
where the $U_i$ follow a standard normal distribution, independent of other components. To evaluate the performance of cross-validation estimators, we generate
\[
x_{ji}=\mu_i+\sqrt k u_{ji}
\]
for $j=1,\ldots, k$, where the $u_{ji}$ are draws from independent standard normal distributions. As a result, the averages
\[
X_{ki}=\frac{1}{k}\sum_{j=1}^k x_{ji}
\]
have the same distributions as the $X_i$ in equation (\ref{equation:Xi}), which makes the comparison of between the cross-validation estimators and the SURE and NPEB estimators a meaninful one.  For cross-validation estimators we consider $k=4, 20$. 

\paragraph{Estimators}
The SURE criterion function employed in the simulations is the one in equation (\ref{equation:surecrit}) where, for the pretest estimator, the density of $X$ is 
estimated with a normal kernel and the bandwidth implied by ``Silverman's rule of thumb''.\footnote{See \cite{silverman1986density} equation (3.31).} The cross-validation criterion function employed in the simulations is a leave-one-out version of the one considered in Section \ref{ssec:CV},
\begin{equation}
\label{equation:loocv}
r_{n,k}(\lambda)=\sum_{j=1}^k \left(\frac{1}{n}\sum_{i=1}^n(m(X_{-ji},\lambda)-x_{ji})^2\right),
\end{equation}
where $X_{-ji}$ is the average of $\{x_{1i},\ldots, x_{ki}\}\setminus x_{ji}$. Notice that because of the result in Theorem \ref{theorem:cvunif} applies to each of the $k$ terms on the right-hand-side of equation (\ref{equation:loocv}) it also applies to $r_{n,k}(\lambda)$ as defined on the left-hand-side of the same equation. The cross validation estimator employed in our simulations is $m(X_{ki},\lambda)$, with $\lambda$ evaluated at the minimizer of (\ref{equation:loocv}). 

\paragraph{Results}
Tables \ref{table:simloss50}, \ref{table:simloss200}, and \ref{table:simloss1000} report average compound risk across 1000 simulations for $n=50$, $n=200$ and $n=1000$, respectively. Each row corresponds to a particular value of $(p,\mu_0,\sigma_0)$, and each column corresponds to a particular estimator/regularization criterion. The results are coded row-by-row on a continuous color scale which varies from dark blue (minimum row value) to light yellow (maximum row value). 

Several clear patterns emerge from the simulation results. First, even for a dimensionality as modest as $n=50$, the patterns in Figure \ref{fig:RiskSpikeNormal}, which were obtained for oracle choices of regularization parameters, are reproduced in Tables \ref{table:simloss50} to \ref{table:simloss1000} for the same estimators but using data-driven choices of regularization parameters. As in Figure \ref{fig:RiskSpikeNormal}, among ridge, lasso and pretest, ridge dominates when there is little or no sparsity in the parameters of interest, pretest dominates when the distribution of non-zero parameters is substantially separated from zero, and lasso dominates in the intermediate cases. Second, while the results in \cite{jiang2009eb} suggest good performance of nonparametric estimators of $\bar m_\pi^*$ for large $n$, the simulation results in Tables \ref{table:simloss50} and \ref{table:simloss200} indicate that the performance of NPEB may be substantially worse than the performance of the other machine learning estimators in the table, for moderate and small $n$. In particular, the performance of the NPEB estimator suffers in the settings with low or no sparsity, especially when the distribution of the non-zero values of $\mu_1,\ldots, \mu_n$ has considerable dispersion. This is explained by the fact that, in practice, the NPEB estimator approximates the distribution of $\mu$ using a discrete distribution supported on a small number of values. When most of the probability mass of the true distribution of $\mu$ is also concentrated around a small number of values (that is, when $p$ is large or $\sigma_0$ is small), the approximation employed by the NPEB estimator is accurate and the performance of the NPEB estimator is good. This is not the case, however, when the true distribution of $\mu$ cannot be closely approximated with a small number of values (that is, when $p$ is small and $\sigma_0$ is large). Lasso shows a remarkable degree of robustness to the value of $(p,\mu_0,\sigma_0)$, which makes it an attractive estimator in practice. For large $n$, as in Table \ref{table:simloss1000}, NPEB dominates except in settings with no sparsity and a large dispersion in $\mu$ ($p=0$ and $\sigma_0$ large).
 
\section{Applications}
\label{sec:applications}

In this section, we apply our results to three data sets from the empirical economics literature. The first application, based on \cite{chetty2015impacts}, estimates the effect of living in a given commuting zone during childhood on intergenerational income mobility.
The second application, based on \cite{dellavigna2010detecting}, estimates changes in the stock prices of arms manufacturers following changes in the intensity of conflicts in countries under arms trade embargoes. The third application uses data from the 2000 census of the US, previously employed in \cite{angrist2006miss} and \cite{belloni2011high}, to estimate a nonparametric Mincer regression equation of log wages on education and potential experience.

For all applications we normalize the observed $X_i$ by their estimated standard error. Note that this normalization (i) defines the implied loss function, which is quadratic error loss for estimation of the normalized latent parameter $\mu_i$, and (ii) defines the class of estimators considered, which are componentwise shrinkage estimators based on the normalized $X_i$.

\subsection{Neighborhood Effects: Chetty and Hendren (2015)}
\label{section:location}

Chetty and Hendren (2015) use information on income at age 26 for individuals who moved between commuting zones during childhood to estimate the effects of location on income. Identification comes from comparing differently aged children of the same parents, who are exposed to different locations for different durations in their youth. In the context of this application, $X_i$ is the (studentized) estimate of the effect of spending an additional year of childhood in commuting zone $i$, conditional on parental income rank, on child income rank relative to the national household income distribution at age 26.\footnote{The data employed in this section were obtained from \url{http://www.equality-of-opportunity.org/images/nbhds_online_data_table3.xlsx}. We focus on the estimates for children with parents at the 25th percentile of the national income distribution among parents with children in the same birth cohort.} In this setting, the point zero has no special role; it is just defined, by normalization, to equal the average of commuting zone effects. We therefore have no reason to expect sparsity, nor the presence of a set of effects well separated from zero. Our discussion in Section \ref{sec:riskfunction} would thus lead us to expect that ridge will perform well, and this is indeed what we find.

\begin{figure}\begin{center}
\caption{Neighborhood Effects: SURE Estimates}
\label{fig:SURElocationfigure}
\includegraphics[width = \textwidth]{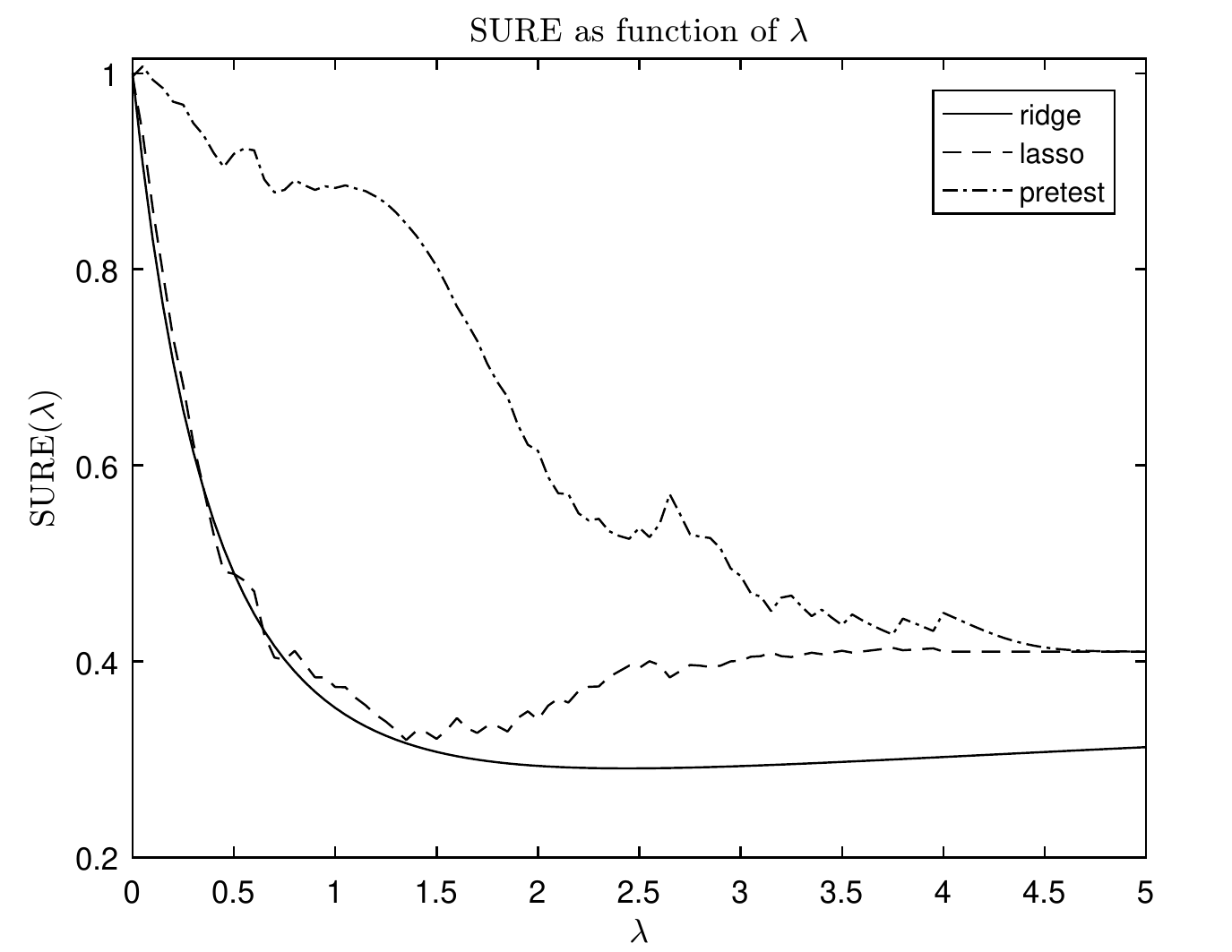}
\end{center}
\end{figure}

Figure \ref{fig:SURElocationfigure} reports SURE estimates of risk for ridge, lasso, and pretest estimators, as functions of $\lambda$. Among the three estimators, minimal estimated risk is equal to 0.29, and it is attained by ridge for $\widehat\lambda_{R,n}=2.44$. Minimal estimated risk for lasso and pretest are 0.31 and 0.41, respectively. The relative performance of the three shrinkage estimators reflects the characteristics of the example and, in particular, the very limited evidence of sparsity in the data.

\begin{figure}\begin{center}
\caption{Neighborhood Effects: Shrinkage Estimators}
\label{fig:estimatorslocation}
\includegraphics[width = \textwidth]{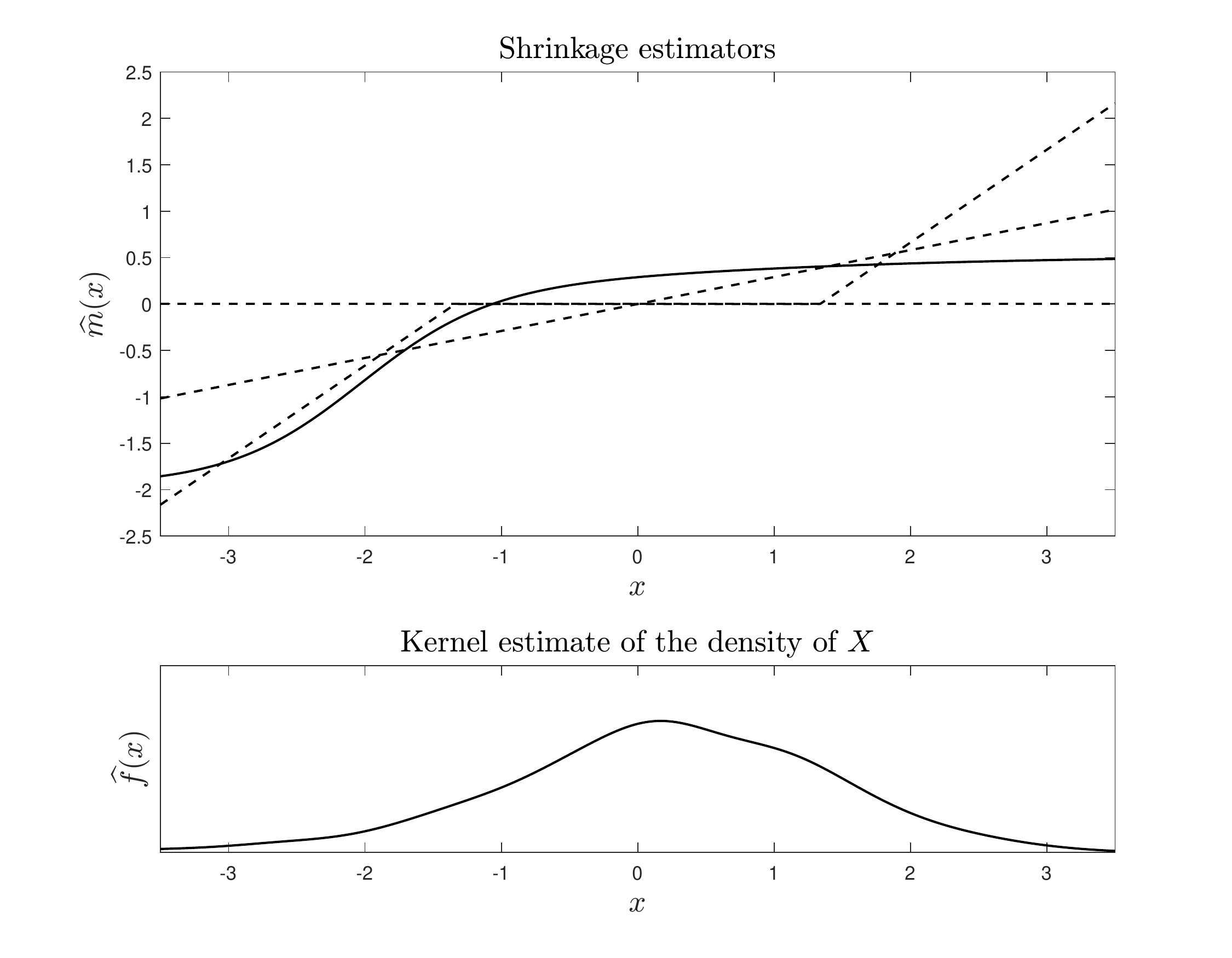}
\end{center}
\floatfoot{\footnotesize The first panel shows the Koenker-Mizera NPEB estimator (solid line) along with the ridge, lasso, and pretest estimators (dashed lines) evaluated at SURE-minimizing values of the regularization parameters. The ridge estimator is linear, with positive slope equal to estimated risk, $0.29$. Lasso is piecewise linear, with kinks at the positive and negative versions of the SURE-minimizing value of the regularization parameter, $\widehat\lambda_{L,n} =  1.34$. Pretest is flat at zero, because SURE is minimized for values of $\lambda$ higher than the maximum absolute value of $X_1,\ldots, X_n$. The second panel shows a kernel estimate of the distribution of $X$.}
\end{figure}

The first panel of Figure \ref{fig:estimatorslocation} shows the Koenker-Mizera NPEB estimator (solid line) along with the ridge, lasso, and pretest estimators (dashed lines) evaluated at SURE-minimizing values of the regularization parameters. The identity of the estimators can be easily recognized from their shape. The ridge estimator is linear, with positive slope equal to estimated risk, $0.29$. Lasso has the familiar piecewise linear shape, with kinks at the positive and negative versions of the SURE-minimizing value of the regularization parameter, $\widehat\lambda_{L,n} =  1.34$. Pretest is flat at zero, because SURE is minimized for values of $\lambda$ higher than the maximum absolute value of $X_1,\ldots, X_n$. The second panel shows a kernel estimate of the distribution of $X$.\footnote{To produce a smooth depiction of densities, for the panels reporting densities in this section we use the normal reference rule to choose the bandwidth. See, e.g., \cite{silverman1986density} equation (3.28).} Among ridge, lasso, and pretest, ridge best approximates the optimal shrinkage estimator over most of the estimated distribution of $X$. Lasso comes a close second, as evidenced in the minimal SURE values for the three estimators, and pretest is way off. Despite substantial shrinkage, these estimates suggest considerable heterogeneity in the effects of childhood neighborhood on earnings. In addition, as expected given the nature of this application, we do not find evidence of sparsity in the location effects estimates. 

\subsection{Detecting Illegal Arms Trade: \cite{dellavigna2010detecting}}

\cite{dellavigna2010detecting} use changes in stocks prices of arms manufacturing companies at the time of large changes in the intensity of conflicts in countries under arms-trade embargoes to detect illegal arms trade. In this section, we apply the estimators in Section \ref{sec:datadrivenregularization} to data from the Della Vigna and La Ferrara study.\footnote{\cite{dellavigna2010detecting} divide their sample of arms manufacturers in two groups, depending on whether the company is head-quartered in a country with a high or low level of corruption. They also divide the events of changes in the intensity of the conflicts in embargo areas in two groups, depending on whether the intensity of the conflict increased or decreased at the time of the event. For concreteness, we use the 214 event study estimates for events of increase in the intensity of conflicts in arms embargo areas and for companies in high-corruption countries. The data for this application is available at \url{http://eml.berkeley.edu/~sdellavi/wp/AEJDataPostingZip.zip}.}

In contrast to the location effects example in Section \ref{section:location}, in this application there are reasons to expect a certain amount of sparsity, if changes in the intensity of the conflicts in arms-embargo areas do not affect the stock prices of arms manufacturers that comply with the embargoes.\footnote{In the words of \cite{dellavigna2010detecting}: ``If a company is not trading or trading legally, an event increasing the hostilities should not affect its stock price or should affect it adversely, since it delays the removal of the embargo and hence the re-establishment of legal sales. Conversely, if a company is trading illegally, the event should increase its stock price, since it increases the demand for illegal weapons.''}
Economic theory would suggest this to be the case if there are fixed costs for violating the embargo.
In this case, our discussion of Section \ref{sec:riskfunction} would lead us to expect that pretest might be optimal, which is again what we find.

\begin{figure}\begin{center}
\caption{Arms Event Study: SURE Estimates}
\label{fig:SUREarmsfigure}
\includegraphics[width = \textwidth]{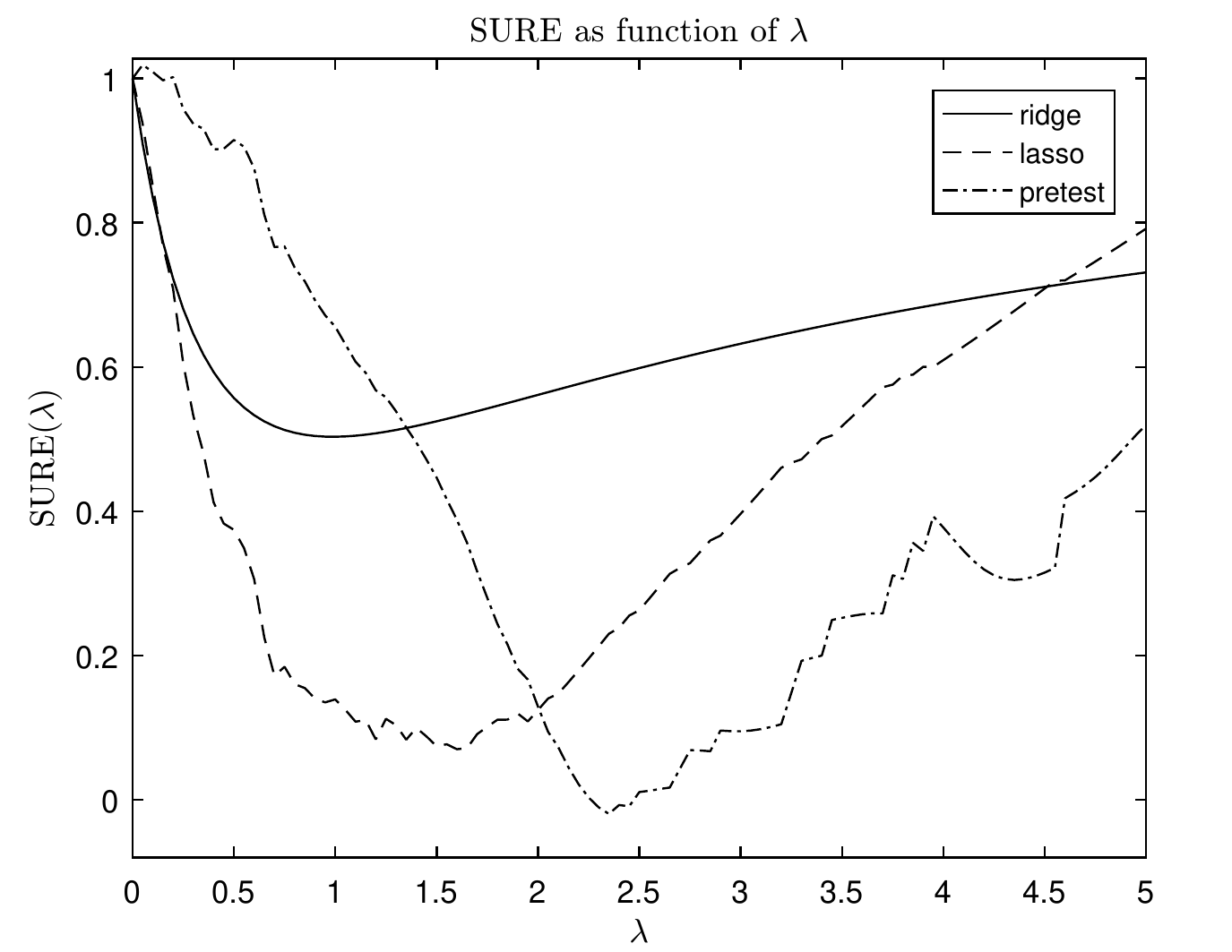}
\end{center}
\end{figure}

Figure \ref{fig:SUREarmsfigure} shows SURE estimates for ridge, lasso, and pretest.
Pretest has the lowest estimated risk, for $\widehat\lambda_{PT,n}=2.39$,\footnote{Notice that the pretest's SURE estimate attains a negative minimum value. This could be a matter of estimation variability, of inappropriate choice of bandwidth for the estimation of the density of $X$ in small samples, or it could reflect misspecification of the model (in particular, Gaussianity of $X$ given $\mu$).} followed by lasso, for $\widehat\lambda_{L,n}=1.50$.

\begin{figure}\begin{center}
\caption{Arms Event Study: Shrinkage Estimators}
\label{fig:estimatorsArms}
\includegraphics[width = \textwidth]{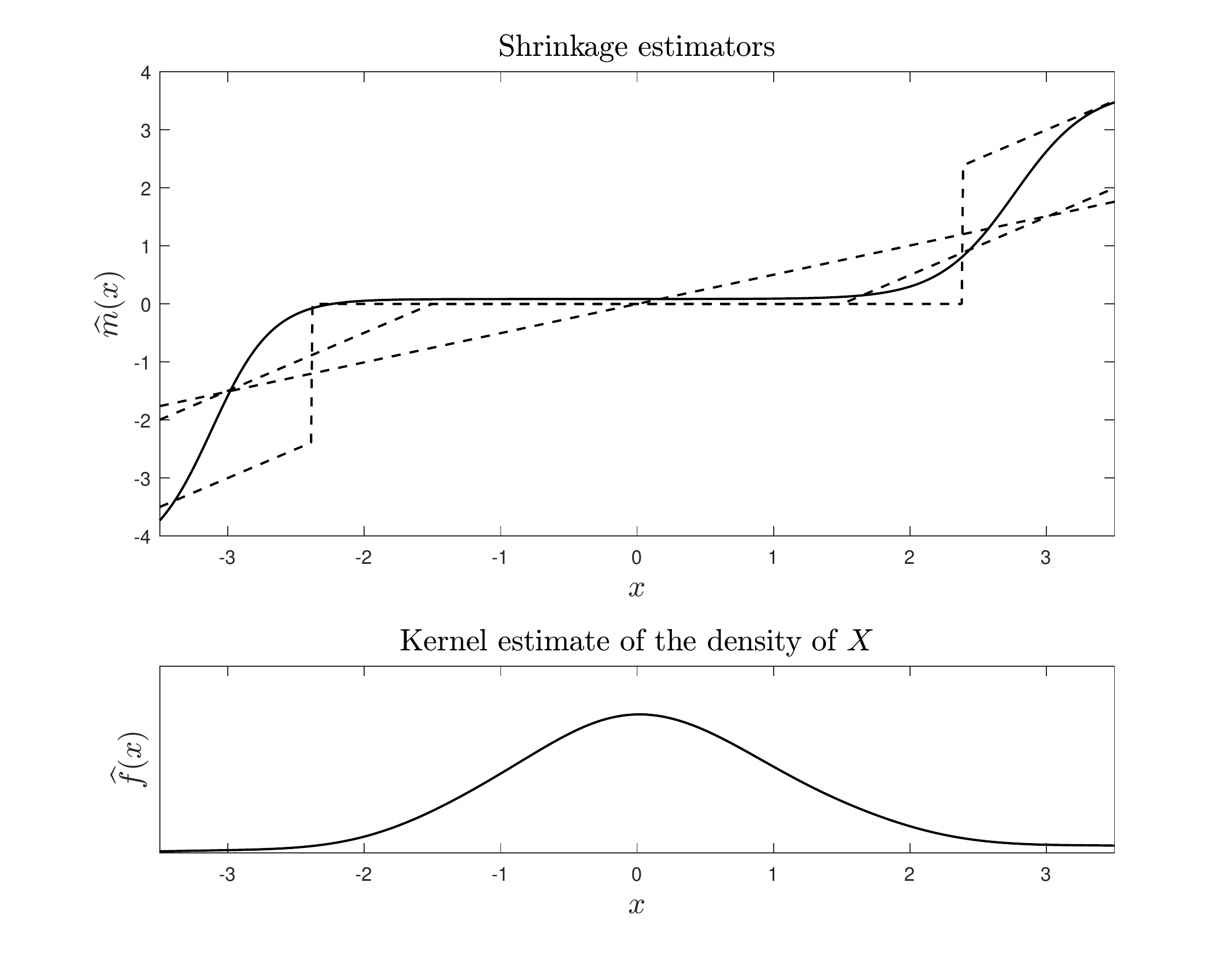}
\end{center}
\floatfoot{\footnotesize The first panel shows the Koenker-Mizera NPEB estimator (solid line) along with the ridge, lasso, and pretest estimators (dashed lines) evaluated at SURE-minimizing values of the regularization parameters. The ridge estimator is linear, with positive slope equal to estimated risk, $0.50$. Lasso is piecewise linear, with kinks at the positive and negative versions of the SURE-minimizing value of the regularization parameter, $\widehat\lambda_{L,n} =  1.50$. Pretest is discontinuous at $\widehat\lambda_{PT,n} =  2.39$ and $-\widehat\lambda_{PT,n} =  -2.39$.}\
\end{figure}

Figure \ref{fig:estimatorsArms} depicts the different shrinkage estimators and shows that lasso and especially pretest closely approximate the NPEB estimator over a large part of the distribution of $X$. The NPEB estimate suggests a substantial amount of sparsity in the distribution of $\mu$. There is, however, a subset of the support of $X$ around $x=3$ where the estimate of the optimal shrinkage function implies only  a small amount of shrinkage.  Given the shapes of the optimal shrinkage function estimate and of the estimate of the distribution of $X$, it is not surprising that the minimal values of SURE in Figure \ref{fig:SUREarmsfigure} for lasso and pretest are considerably lower than for ridge.

\subsection{Nonparametric Mincer equation: \cite{belloni2011high}}

In our third application, we use data from the 2000 US Census in order to estimate a non-parametric regression of log wages on years of education and potential experience, similar to the example considered in \cite{belloni2011high}.\footnote{The data for this application are available at \url{http://economics.mit.edu/files/384}.} We construct a set of 66 regressors by taking a saturated basis of linear splines in education, fully interacted with the terms of a 6-th order polynomial in potential experience. We orthogonalize these regressors and take the coefficients $X_i$ of an OLS regression of log wages on these orthogonalized regressors as our point of departure. We exclude three coefficients of very large magnitude,\footnote{The three excluded coefficients have values, 2938.04 (the intercept), 98.19, and -77.35. The largest absolute value among the included coefficients is -21.06. Most of the included coefficients are small in absolute value. About 40 percent of them have absolute values smaller than one, and about 60 percent of them have absolute value smaller than two.} which results in $n=63$.
In this application, economics provides less intuition as to what distribution of coefficients to expect. Based on functional analysis considerations, \cite{belloni2011high} argue that for plausible families of functions containing the true conditional expectation function, sparse approximations of the coefficients of series regression as induced by the lasso penalty, have low mean squared error.

\begin{figure}\begin{center}
\caption{Nonparametric Mincer Equation: SURE Estimates}
\label{fig:SUREmincerfigure}
\includegraphics[width = \textwidth]{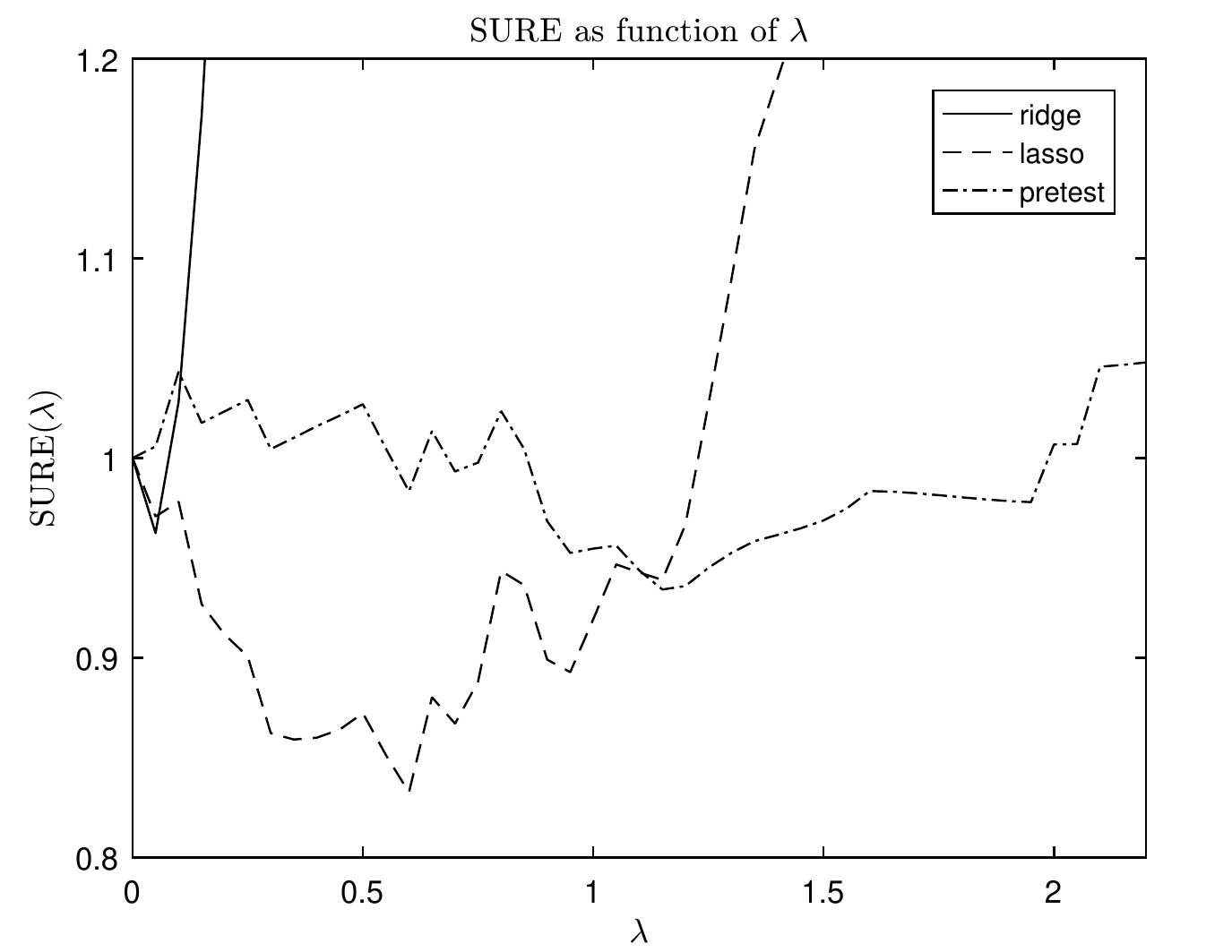}
\end{center}
\end{figure}

Figure \ref{fig:SUREmincerfigure} reports SURE estimates of risk for ridge, lasso and pretest.  In this application, estimated risk for lasso is substantially smaller than for ridge or pretest. 
 
\begin{figure}\begin{center}
\caption{Nonparametric Mincer Equation: Shrinkage Estimators}
\label{fig:estimatorsmincer}
\includegraphics[width = \textwidth]{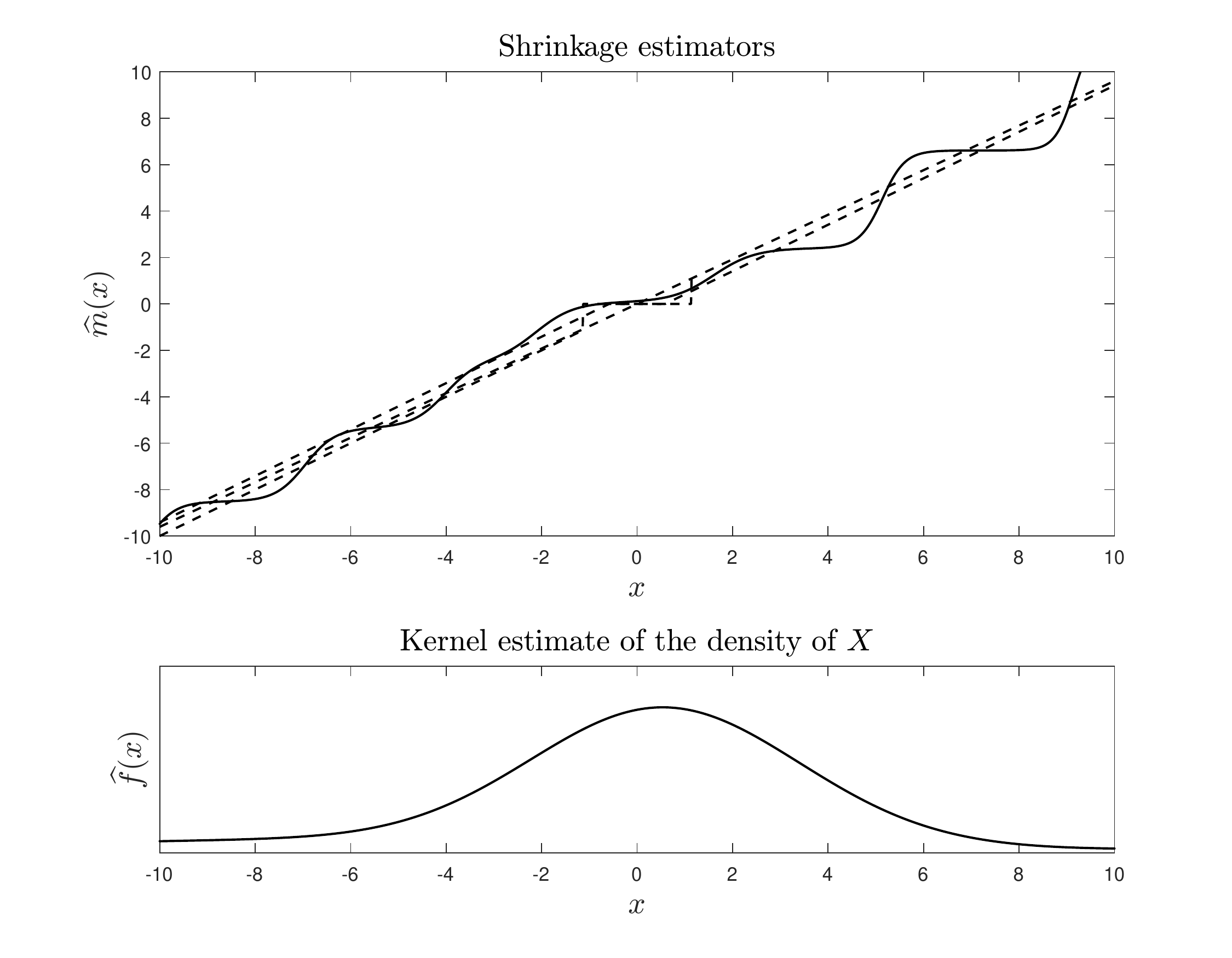}
\end{center}
\floatfoot{\footnotesize The first panel shows the Koenker-Mizera NPEB estimator (solid line) along with the ridge, lasso, and pretest estimators (dashed lines) evaluated at SURE-minimizing values of the regularization parameters. The ridge estimator is linear, with positive slope equal to estimated risk, $0.996$. Lasso is piecewise linear, with kinks at the positive and negative versions of the SURE-minimizing value of the regularization parameter, $\lambda =  0.59$. Pretest is discontinuous at $\widehat\lambda_{PT,n} =  1.14$ and $-\widehat\lambda_{PT,n} =  -1.14$. The second panel shows a kernel estimate of the distribution of $X$.}
\end{figure}

The top panel of Figure \ref{fig:estimatorsmincer} reports the three regularized estimators, ridge, lasso, and pretest, evaluated at the data-driven choice of regularization parameter, along with the Koenker-Mizera NPEB estimator. In order to visualize the differences between the estimates close to the origin, where most of the coefficients are, we report the value of the estimates for $x\in [-10, 10]$. The bottom panel of Figure \ref{fig:estimatorsmincer} reports an estimate of the density of $X$. Locally, the shape of the NPEB estimate looks similar to a step function. This behavior is explained by the fact that the NPEB estimator is based on an approximation to the distribution of $\mu$ that is supported on a finite number of values. However, over the whole range of $x$ in the Figure \ref{fig:estimatorsmincer}, the NPEB estimate is fairly linear. In view of this close-to-linear behavior of NPEB in the $[10,10]$ interval, the very poor risk performance of ridge relative to lasso and pretest, as evidenced in Figure \ref{fig:SUREmincerfigure}, may appear surprising. This is explained by the fact that in this application, some of the values in $X_1, \ldots, X_n$ fall exceedingly far from the origin. Linearly shrinking those values towards zero induces severe loss. As a result, ridge attains minimal risk for a close-to-zero value of the regularization parameter, $\widehat\lambda_{R,n}=0.04$, resulting in negligible shrinkage. Among ridge, lasso, and pretest, minimal estimated risk is attained by lasso for $\widehat\lambda_{L,n} = 0.59$, which shrinks about 24 percent of the regression coefficients all the way to zero. Pretest induces higher sparsity ($\widehat\lambda_{PT,n}=1.14$, shrinking about 49 percent of the coefficients all the way to zero) but does not improve over lasso in terms of risk.

\section{Conclusion}
\label{sec:conclusion}

The interest in adopting machine learning methods in economics is growing rapidly. Two common features of machine learning algorithms are regularization and data-driven choice of regularization parameters. We study the properties of such procedures. We consider, in particular, the problem of estimating many means $\mu_i$ based on observations $X_i$. This problem arises often in economic applications. In such applications, the ``observations'' $X_i$ are usually equal to preliminary least squares coefficient estimates, like fixed effects.

Our goal is to provide guidance for applied researchers on the use of machine learning estimators. Which estimation method should one choose in a given application? And how should one choose regularization parameters? To the extent that researchers care about the squared error of their estimates, procedures are preferable if they have lower mean squared errors than the competitors.

Based on our results, ridge appears to dominate the alternatives considered when the true effects $\mu_i$ are smoothly distributed, and there is no point mass of true zeros. This is likely to be the case in applications where the objects of interests are the effects of many treatments, such as locations or teachers, and applications that estimate effects for many subgroups. Pretest appears to dominate if there are true zeros and non-zero effects are well separated from zero. This happens in economic applications when there are fixed costs for agents who engage in non-zero behavior. Lasso finally dominates for intermediate cases and appears to do well for series regression, in particular.

Regarding the choice of regularization parameters, we prove a series of results which show that data-driven choices are almost optimal (in a uniform sense) for large-dimensional problems. This is the case, in particular, for choices of regularization parameters that minimize Stein's Unbiased Risk Estimate (SURE), when observations are normally distributed, and for Cross Validation (CV), when repeated observations for a given effect are available. Although not explicitly analyzed in this article, equation (\ref{equation:lsof}) suggests a new empirical selector of regularization parameters based on the minimization of the sample mean square discrepancy between $m(X_i,\lambda)$ and NPEB estimates of $\bar m^*_\pi(X_i)$.

There are, of course, some limitations to our analysis. First, we focus on a restricted class of estimators, those which can be written in the componentwise shrinkage form $\widehat{\mu}_i = m(X_i, \widehat{\lambda})$. This covers many estimators of interest for economists, most notably ridge, lasso, and pretest estimation. Many other estimators in the machine learning literature, such as random forests or neural nets, do not have this tractable form. The analysis of the risk properties of such estimators constitutes an interesting avenue of future research.

Finally, we focus on mean square error. This loss function is analytically quite convenient and amenable to tractable results. Other loss functions might be of practical interest, however, and might be studied using numerical methods. In this context, it is also worth emphasizing again that we were focusing on point estimation, where all coefficients $\mu_i$ are simultaneously of interest. This is relevant for many practical applications such as those discussed above. In other cases, however, one might instead be interested in the estimates $\widehat{\mu}_i$ solely as input for a lower-dimensional decision problem, or in (frequentist) testing of hypotheses on the coefficients $\mu_i$. Our analysis of mean squared error does not directly speak to such questions.

\newpage
\appendix
\centerline{\bf\normalsize Appendix}

\section{Relating prediction problems to the normal means model setup}
\label{asection:prediction}

We have introduced our setup in the canonical form of the problem of estimating many means.
Machine learning methods are often discussed in terms of the problem of minimizing out-of-sample prediction error. The two problems are closely related.
Consider the linear prediction model
\begin{equation*}
\label{equation:ols}
Y =\bs W' \bs\beta + \epsilon,
\end{equation*}
where $Y$ is a scalar random variable, $\bs W$ is an $(n\times 1)$ vector of covariates (features), and $\epsilon | \bs W \sim N(0, \sigma^2)$.\footnote{Linearity of the conditional expectation and normality are assumed here for ease of exposition; both could in principle be dropped in an asymptotic version of the following argument.}
The machine learning literature is often concerned with the problem of predicting the value of $Y$ of a draw of $(Y,\bs W)$ using
\[
\widehat Y = \bs W'\widehat{\bs\beta},
\]
where $\widehat{\bs\beta}$ is an estimator of $\bs\beta$ based on $N$ ($N\geq n$) previous independent draws, $(Y_1,\bs W_{\!1}),\ldots, (Y_N,\bs W_{\!N})$, from the distribution of $(Y, \bs W)$, so $\widehat\beta$ is independent of $(Y,\bs W)$. We evaluate out-of-sample predictions based on the squared prediction error,
\[
\tilde L = (\widehat Y - Y) ^2 = \left (\bs W' (\widehat{\bs\beta} - \bs\beta)\right )^2 + \epsilon^2 + 2 \left (\bs W'(\widehat{\bs\beta} - \bs\beta)\right ) \epsilon.\]
Suppose that the features $\bs W$ for prediction are drawn from the empirical distribution of $\bs W_{\!1},\ldots, \bs W_{\!N}$,\footnote{This assumption is again made for convenience, to sidestep asymptotic approximations} and that $Y$ is drawn from the conditional population distribution of $Y$ given $\bs W$.
The expected squared prediction error, $\tilde R=E[\tilde L]$, is then equal to
\[\tilde R
= \tr\Big(\bs\Omega \cdot E[(\widehat{\bs\beta} - \bs\beta)(\widehat{\bs\beta} - \bs\beta)' ] \Big)+E[\epsilon^2],\]
where 
\[\bs\Omega= \frac{1}{N} \sum_{j=1}^N \bs W_j \bs W'_j.\]
In the special case where the components of $\bs W$ are orthonormal in the sample, $\bs\Omega = \bs I_n$, this immediately yields
\[
\tilde R = \sum_{i=1}^n E[(\widehat\beta_i-\beta_i)^2]+E[\epsilon^2],
\]
where $\widehat\beta_i$ and $\beta_i$ are the $i$-th components of $\widehat{\bs\beta}$ and $\bs\beta$, respectively.
In this special case, we thus get that the risk function for out of sample prediction and the mean squared error for coefficient estimation are the same, up to a constant.\\

 More generally, assume that $\bs\Omega$ has full rank, define $\bs V=\bs\Omega^{-1/2} \bs W$, $\bs \mu = \bs\Omega^{1/2} \bs \beta$, and let $\bs X$ be the coefficients of an ordinary least squares regression of $Y_1,\ldots, Y_N$ on $\bs V_{\!1},\ldots, \bs V_{\!N}$.
This change of coordinates yields, conditional on $\bs W_1, \ldots, \bs W_N$,
\[\bs X \sim N\left (\bs \mu, \frac{\sigma^2}{N} \bs I_n\right ),\]
so that the assumptions of our setup regarding $\bs X$ and $\bs \mu$ hold.
Regularized estimators $\widehat{\bs\mu}$ of $\bs \mu$ can be formed by componentwise shrinkage of $\bs X$.
For any estimator $\widehat{\bs\mu}$ of $\bs \mu$ we can furthermore write the corresponding risk for out of sample prediction as
\[\tilde R= E[(\widehat{\bs\mu} - \bs\mu)'(\widehat{\bs\mu} - \bs\mu)]+E[\epsilon^2].
\]

To summarize: After orthogonalizing the regressors for a linear regression problem, the assumptions of the many means setup apply to the vector of ordinary least squares coefficients.
The risk function for out of sample prediction is furthermore the same as the risk function of the many means problem, if we assume the features for prediction are drawn from the empirical distribution of observed features.

\section{Assuming oracle knowledge of zeros is not uniformly valid}
\label{asection:oracle}

Consider the pretest estimator, $m_{PT}(X_i,\widehat\lambda_n)$. An alternative approximation to the risk of the pretest estimator is given by the risk of the infeasible estimator based on oracle-knowledge of true zeros,
\[
m_{PT}^{0,\mu}(X_i) = 1(\mu_i\neq 0) X_i.
\]
As we show now, this approximation is not uniformly valid, which illustrates that uniformity is not a trivial requirement.
Consider the following family $\mathcal Q$  of data generating processes,
\begin{align*}
X|\mu & \sim N(\mu,1),\\
P(\mu=0) &= p,\\
P(\mu=\mu_0) &= 1-p.
\end{align*}
It is easy to check that
\[
\bar R(m_{PT}^{0,\mu}(\cdot),\pi)= 1-p,
\]
for all $\pi\in\mathcal Q$. By Proposition \ref{prop:spikenormal}, for $\pi\in\mathcal Q$, the integrated risk of the pretest estimator is
\begin{align*}
\bar R(m_{PT}(\cdot,\lambda),\pi) &= 2  \Big(\Phi(-\lambda)
+ \lambda\phi (\lambda)\Big)p\\
&+ \Big(1+\Phi(-\lambda-\mu_0) -\Phi(\lambda-\mu_0) + (\Phi(\lambda-\mu_0)
-\Phi(-\lambda-\mu_0)) \mu_0^2\\
&-\phi(\lambda-\mu_0)\big(-\lambda+\mu_0\big) -\phi(-\lambda-\mu_0)\big(-\lambda-\mu_0\big)\Big) (1-p).
\end{align*}
We have shown above that data-driven choices of $\lambda$ are uniformly risk consistent, so their integrated risk is asymptotically equal to $\min_{\lambda\in [0,\infty]} R(m_{PT}(\cdot,\lambda),\pi)$.
It follows that the risk of $m_{PT}^{0,\mu}(\cdot)$ provides a uniformaly valid approximation to the risk of $m_{PT}(\cdot,\widehat\lambda)$ if and only if
\begin{equation}
\label{equation:uniformvalid}
\min_{\lambda\in [0,\infty]} \bar R(m_{PT}(\cdot,\lambda),\pi)= 1-p,\quad \forall \pi\in\mathcal Q.
\end{equation}
It is easy to show that equation (\ref{equation:uniformvalid}) is violated. Consider, for example, $(p,\mu_0) = (1/2,\sqrt 2)$. Then, the minimum value of $\bar R(m_{PT}(,\lambda),\pi)$ is equal to one (achieved at  $\lambda = 0$ and $\lambda=\infty$). Therefore,
\[\min_{\lambda\in [0,\infty]} \bar R(m_{PT}(,\lambda),\pi) = 1 > 0.5 = \bar R(m_{PT}^{0,\mu}(\cdot),\pi).\]
Moreover, equation (\ref{equation:uniformvalid}) is also violated in the opposite direction. Notice that
\[\lim_{\lambda\rightarrow\infty} \bar R(m_{PT}(\cdot,\lambda),\pi)=(1-p)\mu_0^2.\]
As a result, if $|\mu_0|<1$ we obtain
\[
\min_{\lambda\in [0,\infty]} \bar R(m_{PT}(,\lambda),\pi) < 1-p = \bar R(m_{PT}^{0,\mu}(\cdot),\pi),
\]
which violates equation (\ref{equation:uniformvalid}).

\section{Proofs}

\noindent\textbf{Proof of Theorem \ref{theo:mstar}:}
\begin{align*}
R_n(m(.,\lambda) , \bs P) &=  \frac{1}{n} \sum_{i=1}^n E[(m(X_i,\lambda)-\mu_i)^2|P_i]\\
&=E\big[(m(X_I,\lambda) - \mu_I)^2|\bs P\big]\\
&=  E\big[E[(m^*_{\bs P}(X_I) - \mu_I)^2 | X_I,\bs P]|\bs P\big]  + E\big[(m(X_I,\lambda) - m^*_{\bs P}(X_I))^2|\bs P\big]\\
&= v^*_{\bs P} + E\big[(m(X_I,\lambda) - m^*_{\bs P}(X_I))^2|\bs P\big].
\end{align*}
The second equality in this proof is termed {\it the fundamental theorem of compound decisions} in \cite{jiang2009eb}, who credit \cite{robbins1951submin}. Finiteness of $\mu_1,\ldots, \mu_n$, and $\sup_{\lambda\in[0,\infty]} E[(m(X_I,\lambda))^2|{\bs P}]$ implies that all relevant expectations are finite.\quelle{$\square$}

\noindent\textbf{Proof of Lemma \ref{lem:componentwise}:} Notice that
\[
m_R(x,\lambda)-\mu_i = \left(\frac{1}{1+\lambda}\right)(x-\mu_i)-\left(\frac{\lambda}{1+\lambda}\right)\mu_i.
\]
The result for ridge equals the second moment of this expression. For pretest, notice that
\[
m_{PT}(x,\lambda)-\mu_i = 1(|x|>\lambda) (x-\mu_i)-1(|x|\leq\lambda)\mu_i.
\]
Therefore,
\begin{equation}
\label{equation:PTcomprisk1}
R(m_{PT}(\cdot,\lambda),P_i)= E\big[(X_i-\mu_i)^21(|X_i|>\lambda)\big]+ \mu_i^2 \Pr\big(|X_i|\leq\lambda\big).
\end{equation}
Using the fact that $\phi'(v)=-v\phi(v)$ and integrating by parts, we obtain
\begin{align*}
\int_a^b v^2 \phi(v)\, dv&=\int_a^b \phi(v)\, dv-\Big[b\phi(b)-a\phi(a)\Big]\\
&=\Big[\Phi(b)-\Phi(a)\Big]-\Big[b\phi(b)-a\phi(a)\Big].
\end{align*}
Now,
\begin{align}
\label{equation:PTcomprisk2}
E\big[(X_i-\mu_i)^21(|X_i|>\lambda)\big]&=\sigma_i^2E\Bigg[\Bigg(\frac{X_i-\mu_i}{\sigma_i}\Bigg)^21(|X_i|>\lambda)\Bigg]\nonumber\\
&=\Bigg(1 + \Phi\Big(\displaystyle\frac{-\lambda-\mu_i}{\sigma_i}\Big)-\Phi\Big(\displaystyle\frac{\lambda-\mu_i}{\sigma_i}\Big) \Bigg)\sigma_i^2\nonumber\\ &+\Bigg(\Big(\displaystyle\frac{\lambda-\mu_i}{\sigma_i}\Big)\phi\Big(\displaystyle\frac{\lambda-\mu_i}{\sigma_i}\Big)
-\Big(\displaystyle\frac{-\lambda-\mu_i}{\sigma_i}\Big)\phi\Big(\displaystyle\frac{-\lambda-\mu_i}{\sigma_i}\Big)\Bigg)\sigma_i^2.
\end{align}
The result for the pretest estimator now follows easily from equations (\ref{equation:PTcomprisk1}) and (\ref{equation:PTcomprisk2}). For lasso, notice that
\begin{align*}
m_L(x,\lambda)-\mu_i&= 1(x<-\lambda)(x+\lambda-\mu_i)+1(x>\lambda)(x-\lambda-\mu_i)-1(|x|\leq\lambda)\mu_i \\
&=1(|x|>\lambda)(x-\mu_i)+(1(x<-\lambda)-1(x>\lambda))\lambda-1(|x|\leq\lambda)\mu_i .
\end{align*}
Therefore,
\begin{align}
\label{equation:Lcomprisk1}
R(m_L(\cdot,\lambda),P_i)&=E\big[(X_i-\mu_i)^21(|X_i|>\lambda)\big]+\lambda^2E[1(|X_i|>\lambda)]+\mu_i^2E[1(|X_i| \leq \lambda)]\nonumber\\
&+2\lambda\Big(E\big[(X_i-\mu_i)1(X_i<-\lambda)\big]-E\big[(X_i-\mu_i)1(X_i>\lambda)\big]\Big)\nonumber\\
&= R(m_{PT}(\cdot,\lambda),P_i) + \lambda^2E[1(|X_i|>\lambda)]\nonumber\\
&+2\lambda\Big(E\big[(X_i-\mu_i)1(X_i<-\lambda)\big]-E\big[(X_i-\mu_i)1(X_i>\lambda)\big]\Big).
\end{align}
Notice that
\[
\int_a^b v \phi(v) dv = \phi(a)-\phi(b).
\]
As a result,
\begin{equation}
\label{equation:Lcomprisk2}
E\big[(X_i-\mu_i)1(X_i<-\lambda)\big]-E\big[(X_i-\mu_i)1(X_i>\lambda)\big]=-\sigma_i\left(\phi\Big(\displaystyle\frac{-\lambda-\mu_i}{\sigma_i}\Big)
+\phi\Big(\displaystyle\frac{\lambda-\mu_i}{\sigma_i}\Big)\right).
\end{equation}
Now, the result for lasso follows from equations (\ref{equation:Lcomprisk1}) and (\ref{equation:Lcomprisk2}).
\hfill$\square$

\noindent\textbf{Proof of Proposition \ref{prop:spikenormal}:}
The results for ridge are trivial. For lasso, first notice that the integrated risk at zero is:
\[
R_{0}(m_{L}(\cdot,\lambda),\pi)=2\Phi\Big(\displaystyle\frac{-\lambda}{\sigma}\Big)(\sigma^2+\lambda^2)
-2\Big(\displaystyle\frac{\lambda}{\sigma}\Big)\phi\Big(\displaystyle\frac{\lambda}{\sigma}\Big)\sigma^2.
\]
Next, notice that
\[
\int \Phi\Big(\displaystyle\frac{-\lambda-\mu}{\sigma}\Big)\frac{1}{\sigma_0}\phi\Big(\displaystyle\frac{\mu_0-\mu}{\sigma_0}\Big)d\mu
= \Phi\Bigg(\displaystyle\frac{-\lambda-\mu_0}{\sqrt{\sigma_0^2+\sigma^2}}\Bigg),
\]
\[
\int \Phi\Big(\displaystyle\frac{\lambda-\mu}{\sigma}\Big)\frac{1}{\sigma_0}\phi\Big(\displaystyle\frac{\mu_0-\mu}{\sigma_0}\Big)d\mu
= \Phi\Bigg(\displaystyle\frac{\lambda-\mu_0}{\sqrt{\sigma_0^2+\sigma^2}}\Bigg),
\]
\[
\int \Big(\displaystyle\frac{-\lambda-\mu}{\sigma}\Big)\phi\Big(\displaystyle\frac{\lambda-\mu}{\sigma}\Big)  \frac{1}{\sigma_0}\phi\Big(\displaystyle\frac{\mu_0-\mu}{\sigma_0}\Big)d\mu
=-\left(\frac{1}{\sqrt{\sigma_0^2+\sigma^2}}\phi\Big(\displaystyle\frac{\lambda-\mu_0}{\sqrt{\sigma_0^2+\sigma^2}}\Big)\right)
\left(\lambda+\frac{\mu_0\sigma^2+\lambda\sigma_0^2}{\sigma_0^2+\sigma^2}\right)
\]
\[
\int \Big(\displaystyle\frac{-\lambda+\mu}{\sigma}\Big)\phi\Big(\displaystyle\frac{-\lambda-\mu}{\sigma}\Big)  \frac{1}{\sigma_0}\phi\Big(\displaystyle\frac{\mu_0-\mu}{\sigma_0}\Big)d\mu
=-\left(\frac{1}{\sqrt{\sigma_0^2+\sigma^2}}\phi\Big(\displaystyle\frac{-\lambda-\mu_0}{\sqrt{\sigma_0^2+\sigma^2}}\Big)\right)
\left(\lambda-\frac{\mu_0\sigma^2-\lambda\sigma_0^2}{\sigma_0^2+\sigma^2}\right).
\]
The integrals involving $\mu^2$ are more involved. Let $v$ be a Standard normal variable independent of $\mu$. Notice that,
\begin{align*}
\int \mu^2 \Phi\Big(\displaystyle\frac{\lambda-\mu}{\sigma}\Big) \frac{1}{\sigma_0}\phi\Big(\displaystyle\frac{\mu-\mu_0}{\sigma_0}\Big)d\mu
&=\int \mu^2 \Big(\int I_{[v\leq (\lambda-\mu)/\sigma]}\phi(v)dv\Big) \frac{1}{\sigma_0}\phi\Big(\displaystyle\frac{\mu-\mu_0}{\sigma_0}\Big) d\mu\\
&=\int \Big(\int \mu^2 I_{[\mu\leq \lambda-\sigma v]}\frac{1}{\sigma_0}\phi\Big(\displaystyle\frac{\mu-\mu_0}{\sigma_0}\Big) d\mu\Big) \phi(v)dv.
\end{align*}
Using the change of variable $u=(\mu-\mu_0)/\sigma_0$, we obtain,
\begin{align*}
\int \mu^2 I_{[\mu\leq \lambda-\sigma v]}\frac{1}{\sigma_0}\phi\Big(\displaystyle\frac{\mu-\mu_0}{\sigma_0}\Big) d\mu
&=\int (\mu_0+\sigma_0 u)^2 I_{[u\leq (\lambda-\mu_0-\sigma v)/\sigma_0]}\phi(u) du\\
&=\Phi\Big(\displaystyle\frac{\lambda-\mu_0-\sigma v}{\sigma_0}\Big)\mu_0^2-2\phi\Big(\displaystyle\frac{\lambda-\mu_0-\sigma v}{\sigma_0}\Big)\sigma_0\mu_0\\
&+\Bigg(\Phi\Big(\displaystyle\frac{\lambda-\mu_0-\sigma v}{\sigma_0}\Big)-\Big(\displaystyle\frac{\lambda-\mu_0-\sigma v}{\sigma_0}\Big)\phi\Big(\displaystyle\frac{\lambda-\mu_0-\sigma v}{\sigma_0}\Big)\Bigg)\sigma_0^2\\
&=\Phi\Big(\displaystyle\frac{\lambda-\mu_0-\sigma v}{\sigma_0}\Big)(\mu_0^2+\sigma_0^2)-\phi\Big(\displaystyle\frac{\lambda-\mu_0-\sigma v}{\sigma_0}\Big)\sigma_0(\lambda+\mu_0-\sigma v).
\end{align*}
Therefore,
\begin{align*}
\int \mu^2 \Phi\Big(\displaystyle\frac{\lambda-\mu}{\sigma}\Big) \frac{1}{\sigma_0}\phi\Big(\displaystyle\frac{\mu-\mu_0}{\sigma_0}\Big)d\mu&=
\Phi\Bigg(\displaystyle\frac{\lambda-\mu_0}{\sqrt{\sigma_0^2+\sigma^2}}\Bigg)(\mu_0^2+\sigma_0^2)\\
&-\frac{1}{\sqrt{\sigma_0^2+\sigma^2}}\phi\Bigg(\displaystyle\frac{\lambda-\mu_0}{\sqrt{\sigma_0^2+\sigma^2}}\Bigg)(\lambda+\mu_0)\sigma_0^2\\
&+\sigma_0^2\sigma^2 \frac{1}{\sqrt{\sigma_0^2+\sigma^2}}\phi\Bigg(\displaystyle\frac{\lambda-\mu_0}{\sqrt{\sigma_0^2+\sigma^2}}\Bigg)
\Bigg(\displaystyle\frac{\lambda-\mu_0}{\sigma_0^2+\sigma^2}\Bigg).
\end{align*}
Similarly,
\begin{align*}
\int \mu^2 \Phi\Big(\displaystyle\frac{-\lambda-\mu}{\sigma}\Big) \frac{1}{\sigma_0}\phi\Big(\displaystyle\frac{\mu-\mu_0}{\sigma_0}\Big)d\mu
&= \Phi\Bigg(\displaystyle\frac{-\lambda-\mu_0}{\sqrt{\sigma_0^2+\sigma^2}}\Bigg)(\mu_0^2+\sigma_0^2)\\
&-\frac{1}{\sqrt{\sigma_0^2+\sigma^2}}\phi\Bigg(\displaystyle\frac{-\lambda-\mu_0}{\sqrt{\sigma_0^2+\sigma^2}}\Bigg)(-\lambda+\mu_0)\sigma_0^2\\
&+\sigma_0^2\sigma^2 \frac{1}{\sqrt{\sigma_0^2+\sigma^2}}\phi\Bigg(\displaystyle\frac{-\lambda-\mu_0}{\sqrt{\sigma_0^2+\sigma^2}}\Bigg)
\Bigg(\displaystyle\frac{-\lambda-\mu_0}{\sigma_0^2+\sigma^2}\Bigg).
\end{align*}
The integrated risk conditional on $\mu\neq 0$ is
\begin{align*}
R_{1}(m_{L}(\cdot,\lambda),\pi)=\Bigg(&1+\Phi\Bigg(\displaystyle\frac{-\lambda-\mu_0}{\sqrt{\sigma_0^2+\sigma^2}}\Bigg)
-\Phi\Bigg(\displaystyle\frac{\lambda-\mu_0}{\sqrt{\sigma_0^2+\sigma^2}}\Bigg)\Bigg)(\sigma^2+\lambda^2)\\
&+ \Bigg(\Phi\Bigg(\displaystyle\frac{\lambda-\mu_0}{\sqrt{\sigma_0^2+\sigma^2}}\Bigg)
-\Phi\Bigg(\displaystyle\frac{-\lambda-\mu_0}{\sqrt{\sigma_0^2+\sigma^2}}\Bigg)\Bigg)(\mu_0^2+\sigma_0^2)\\
&-\frac{1}{\sqrt{\sigma_0^2+\sigma^2}}\phi\Bigg(\displaystyle\frac{\lambda-\mu_0}{\sqrt{\sigma_0^2+\sigma^2}}\Bigg)(\lambda+\mu_0)(\sigma_0^2+\sigma^2)\\
&-\frac{1}{\sqrt{\sigma_0^2+\sigma^2}}\phi\Bigg(\displaystyle\frac{-\lambda-\mu_0}{\sqrt{\sigma_0^2+\sigma^2}}\Bigg)(\lambda-\mu_0)(\sigma_0^2+\sigma^2).
\end{align*}
The results for pretest follow from similar calculations.
\hfill$\square$

The next lemma is used in the proof of Theorem \ref{theorem:loss}.
\begin{lemma}
\label{lemma:infsup}
For any two real-valued functions, $f$ and $g$,
\[
\Big|\inf f-\inf g\Big|\leq \sup |f-g|.
\]
\end{lemma}

{\bf Proof:} The result of the lemma follows directly from
\begin{align*}
\inf f \geq \inf g - \sup |f-g|,\\
\intertext{and}
\inf g \geq \inf f - \sup |f-g|.
\end{align*}
\hfill$\square$

{\bf Proof of Theorem \ref{theorem:loss}:} Because $\bar v_\pi$ does not depend on $\lambda$, we obtain
\begin{align*}
\Big(L_n(\lambda)-L_n(\widehat\lambda_n)\Big)-\Big(r_n(\lambda)-r_n(\widehat\lambda_n)\Big)
&=\Big(L_n(\lambda)-\bar R_\pi(\lambda)\Big)-\Big(L_n(\widehat\lambda_n)-\bar R_\pi(\widehat\lambda_n)\Big)\\
&+\Big(\bar r_\pi(\lambda)-r_n(\lambda)\Big)-\Big(\bar r_\pi(\widehat\lambda_n)-r_n(\widehat\lambda_n)\Big).
\end{align*}
Applying Lemma \ref{lemma:infsup} we obtain
\begin{align*}
\Big|\Big(\inf_{\lambda\in [0,\infty]}L_n(\lambda)-L_n(\widehat\lambda_n)\Big)-\Big(\inf_{\lambda\in [0,\infty]}r_n(\lambda)-r_n(\widehat\lambda_n)\Big)\Big|
&\leq 2\sup_{\lambda\in [0,\infty]}\Big|L_n(\lambda)-\bar R_\pi(\lambda)\Big|\\
&+2\sup_{\lambda\in [0,\infty]}\Big|\bar r_\pi(\lambda)-r_n(\lambda)\Big|.
\end{align*}
Given that $\widehat\lambda_n$ is the value of $\lambda$ at which $r_n(\lambda)$ attains its minimum, the result of the theorem follows.\hfill$\square$

The following preliminary lemma will be used in the proof of Theorem \ref{theorem:l2conv}.
\begin{lemma}
\label{lem:GCuniform}
For any finite set of regularization parameters, $0=\lambda_0 < \ldots < \lambda_k=\infty$, let
\begin{align*}
u_j&=\sup_{\lambda \in [\lambda_{j-1}, \lambda_{j}]} L(\lambda)\\
l_j&=\inf_{\lambda \in [\lambda_{j-1}, \lambda_{j}]} L(\lambda),
\end{align*}
where $L(\lambda)=(\mu-m(X,\lambda))^2$. Suppose that for any $\epsilon>0$ there is a finite set of regularization parameters, $0=\lambda_0 < \ldots < \lambda_k=\infty$ (where $k$ may depend on $\epsilon$), such that
\begin{equation}
\label{equation:bounddiffs}
\sup_{\pi\in\mathcal Q} \max_{1\leq j\leq k} E_\pi [u_j-l_j] \leq \epsilon
\end{equation}
and
\begin{equation}
\label{equation:boundvars}
\sup_{\pi\in\mathcal Q} \max_{1\leq j\leq k} \max \{\mbox{\em var}_\pi (l_j), \mbox{\em var}_\pi (u_j)\} < \infty.
\end{equation}
Then, equation (\ref{equation:l2conv}) holds.
\end{lemma}

{\bf Proof:} We will use $E_n$ to indicate averages over $(\mu_1,X_1),\ldots , (\mu_n,X_n)$. Let $\lambda \in  [\lambda_{j-1}, \lambda_{j}]$. By construction
\begin{align*}
E_n[L(\lambda)] - E_\pi[L(\lambda)] &\leq E_n[u_j] - E_\pi[l_j] \leq E_n[u_j] - E_\pi[u_j] + E_\pi[u_j-l_j]\\
E_n[L(\lambda)] - E_\pi[L(\lambda)] &\geq E_n[l_j] - E_\pi[u_j] \geq E_n[l_j] - E_\pi[l_j] - E_\pi[u_j-l_j]
\end{align*}
and thus
\begin{align*}
\sup_{\lambda\in[0,\infty]}(E_n[L(\lambda)] -& E_\pi[L(\lambda)])^2 \\ &\leq \max_{1\leq j\leq k} \max \{(E_n[u_j] - E_\pi[u_j])^2, (E_n[l_j] - E_\pi[l_j])^2  \} + \Big(\max_{1\leq j\leq k} E_\pi [u_j-l_j]\Big)^2\\
&+2 \max_{1\leq j\leq k} \max \{|E_n[u_j] - E_\pi[u_j]|, |E_n[l_j] - E_\pi[l_j]|  \}\max_{1\leq j\leq k} E_\pi [u_j-l_j]\\
&\leq \sum_{j=1}^k \Big((E_n[u_j] - E_\pi[u_j])^2+ (E_n[l_j] - E_\pi[l_j])^2 \Big) + \epsilon^2\\
&+2\epsilon\sum_{j=1}^k \Big(|E_n[u_j] - E_\pi[u_j]|+ |E_n[l_j] - E_\pi[l_j]| \Big).
\end{align*}
Therefore,
\begin{align*}
E_\pi\Big[\sup_{\lambda\in[0,\infty]}&(E_n[L(\lambda)] - E_\pi[L(\lambda)])^2\Big]\\
&\leq \sum_{j=1}^k \Big( E_\pi[(E_n[u_j] - E_\pi[u_j])^2]+ E_\pi[E_n[l_j] - E_\pi[l_j])^2] \Big) + \epsilon^2\\
&+2\epsilon\sum_{j=1}^k E_\pi[|E_n[u_j] - E_\pi[u_j]|+ |E_n[l_j] - E_\pi[l_j]|]\\
&\leq \sum_{j=1}^k \Big( \mbox{var}_\pi(u_j)/n+ \mbox{var}_\pi(l_j)/n \Big) + \epsilon^2\\
&+2\epsilon\sum_{j=1}^k \Big(\sqrt{\mbox{var}_\pi(u_j)/n}+ \sqrt{\mbox{var}_\pi(l_j)/n}\Big).
\end{align*}
Now, the result of the lemma follows from the assumption of uniformly bounded variances.\hfill$\square$

{\bf Proof of Theorem \ref{theorem:l2conv}:}
We will show that the conditions of the theorem imply equations (\ref{equation:bounddiffs}) and (\ref{equation:boundvars}) and, therefore, the uniform convergence result in equation (\ref{equation:l2conv}).
Using conditions \ref{item:monotonic} and \ref{item:limits}, along with the convexity of 4th powers, we immediately get bounded variances. Because the maximum of a convex function is achieved at the boundary,
\begin{align*}
\mbox{var}_\pi(u_j) &\leq E_\pi[u_j^2] \leq
E_\pi[\max\{(X-\mu)^4, \mu^4\}]\leq E_\pi[(X-\mu)^4] + E_\pi[\mu^4].
\end{align*}
Notice also that
\[
\mbox{var}_\pi(l_j)\leq E_\pi[l_j^2] \leq E_\pi[u_j^2].
\]
Now, condition \ref{item:moments} implies equation (\ref{equation:boundvars}) in Lemma \ref{lem:GCuniform}.

It remains to find a set of regularization parameters such that $E_\pi[u_j-l_j] <\epsilon$ for all $j$.
Using again the monotonicity of $m(X,\lambda)$ in $\lambda$ and convexity of the square function, we have that the supremum defining $u_j$ is achieved at the boundary,
\[u_j=\max \{L(\lambda_{j-1}), L(\lambda_{j})\},\]
while
\[l_j = \min \{L(\lambda_{j-1}), L(\lambda_{j})\}\]
if $\mu\notin [m(X, \lambda_{j-1}), m(X, \lambda_{j}) ]$ and $l_j=0$, otherwise.
In the former case,
\[u_j- l_j = | L(\lambda_{j}) -L(\lambda_{j-1})|, \]
and in the latter case, $u_j-l_j = \max \{L(\lambda_{j-1}), L(\lambda_{j})\}$. Consider first the case of $\mu\notin [m(X, \lambda_{j-1}), m(X, \lambda_{j}) ]$.
Using the formula $a^2-b^2 = (a+b)(a-b)$ and the shorthand $m_j=m(X,\lambda_j)$, we obtain
\begin{align*}
u_j- l_j &= \left|(m_j-\mu)^2-(m_{j-1}-\mu)^2\right|\\
&=\left|\big((m_j-\mu)+(m_{j-1}-\mu)\big)\big(m_j-m_{j-1}\big)\right|\\
&\leq (|m_j - \mu| +  |m_{j-1} - \mu|) |m_j - m_{j-1}|.
\end{align*}
To check that the same bound applies to the case $\mu\in [m(X, \lambda_{j-1}), m(X, \lambda_{j}) ]$, notice that
\begin{align*}
\max\left\{|m_j-\mu|,|m_{j-1}-\mu|\right\}&\leq |m_j-\mu|+|m_{j-1}-\mu|\intertext{and because $\mu\in [m(X, \lambda_{j-1}), m(X, \lambda_{j}) ]$,}
\max\left\{|m_j-\mu|,|m_{j-1}-\mu|\right\}&\leq |m_j-m_{j-1}|.
\end{align*}
Monotonicity, boundary conditions, and the convexity of absolute values allow one to bound further,
\[u_j-l_j \leq 2 (|X - \mu| +  |\mu|) |m_j - m_{j-1}|.\]
Now, condition \ref{item:regparam} in Theorem \ref{theorem:l2conv} implies equation (\ref{equation:bounddiffs}) in Lemma \ref{lem:GCuniform} and, therefore, the result of the theorem.\hfill$\square$\

{\bf Proof of Lemma \ref{lemma:l2rlp}:}
Conditions 1 and 2 of Theorem \ref{theorem:l2conv} are easily verified to hold for ridge, lasso, and the pretest estimator.
Let us thus discuss condition 4.

Let $\Delta m_j = m(X,\lambda_j) - m(X,\lambda_{j-1})$, and $\Delta \lambda_j = \lambda_j-\lambda_{j-1}$. For ridge, $\Delta m_j$ is given by
\[\Delta m_j = \left (\frac{1}{1+\lambda_j} -\dfrac{1}{1+\lambda_{j-1}} \right ) X\]
so that the requirement follows from finite variances if we choose a finite set of regularization parameters such that
\[\left |\frac{1}{1+\lambda_j} -\dfrac{1}{1+\lambda_{j-1}} \right | \sup_{\pi\in\mathcal Q} E\big[(|X-\mu|+|\mu|)|X|\big]<\epsilon  \]
for all $j=1,\ldots, k$, which is possible by the uniformly bounded moments condition.

For lasso, notice that $|\Delta m_k| = (|X|-\lambda_{k-1})\,1(|X|>\lambda_{k-1})\leq |X|\,1(|X|>\lambda_{k-1})$, and $|\Delta m_j| \leq \Delta \lambda_j$ for $j=1,\ldots, k-1$. We will first verify that
for any $\epsilon>0$ there is a finite $\lambda_{k-1}$ such that condition 4 of the lemma holds for $j=k$. Notice that for any pair of non-negative random variables $(\xi,\zeta)$ such that $E[\xi\,\zeta]<\infty$ and for any positive constant, $c$, we have that
\begin{align*}
E[\xi\zeta]&\geq E[\xi\zeta\, 1(\zeta> c)]\geq c E[\xi 1(\zeta> c)]
\end{align*}
and, therefore,
\[
E[\xi 1(\zeta> c)]\leq \frac{E[\xi\zeta]}{c}.
\]
As a consequence of this inequality, and because $\sup_{\pi\in\mathcal Q}E_\pi[(|X-\mu|+|\mu|)|X|^2]<\infty$ (implied by condition 3), then for any $\epsilon>0$ there exists a finite positive constant, $\lambda_{k-1}$ such that condition 4 of the lemma holds for $j=k$. Given that $\lambda_{k-1}$ is finite, $\sup_{\pi\in\mathcal Q}E_\pi[|X-\mu|+|\mu|]<\infty$ and $|\Delta m_j| \leq \Delta \lambda_j$ imply condition 4 for $j=1,\ldots, k-1$.

For pretest,
\[|\Delta m_j| = 	|X|\, 1(|X| \in (\lambda_{j-1}, \lambda_j]),\]
so that we require that for any $\epsilon>0$ we can find a finite number of regularization parameters, $0=\lambda_0<\lambda_1<\ldots <\lambda_{k-1}<\lambda_k=\infty$, such that
\[E_\pi[(|X - \mu| +  |\mu|) |X|\,1 (|X| \in (\lambda_{j-1}, \lambda_j])]<  \epsilon, \]
for $j=1,\ldots, k$. Applying the Cauchy-Schwarz inequality and uniform boundedness of fourth moments, this condition is satisfied if we can choose uniformly bounded $P_\pi(|X| \in (\lambda_{j-1}, \lambda_j])$, which is possible under the assumption that $X$ is continuously distributed with a (version of the) density that is uniformly bounded.\hfill$\square$

\textbf{Proof of Corollary \ref{corollary:risk}:}
From Theorem \ref{theorem:loss} and Lemma \ref{lemma:infsup}, it follows immediately that
\[
\sup_{\pi\in\mathcal Q} P_\pi\left(\left| L_n(\widehat\lambda_n)-\inf_{\lambda\in [0,\infty]} \bar R_\pi (\lambda)\right|>\epsilon\right)\rightarrow 0.
\]
By definition,
\[\bar{R}(m(., \widehat{\lambda}_n), \pi) = E_\pi[L_n(\widehat\lambda_n)].\]
Equation  \eqref{equation:unifRpicons} thus follows if we can strengthen uniform convergence in probability to uniform $L^1$ convergence.
To do so, we need to show uniform integrability of $L_n(\widehat\lambda_n)$, as per Theorem 2.20 in \cite{van1998asymptotic}.

Monotonicity, convexity of loss, and boundary conditions imply
\[L_n(\widehat\lambda_n) \leq \frac{1}{n}\sum_{i=1}^n \Big(\mu_i^2 + (X_i-\mu_i)^2\Big).\]
Uniform integrability along arbitrary sequences $\pi_n$, and thus $L^1$ convergence, follows from the assumed bounds on moments.
\hfill$\square$

\textbf{Proof of Lemma \ref{lem:SURE}:}
Recall the definition $\bar R(m(.), \pi) = E_\pi[(m(X) - \mu)^2]$.
Expanding the square yields
\begin{align*}
E_\pi[(m(X) - \mu)^2] &= E_\pi[(m(X) - X + X - \mu)^2]\\
&= E_\pi[(X - \mu)^2] + E_\pi[(m(X) - X)^2]+ 2 E_\pi[(X - \mu) (m(X)-X)].
\end{align*}
By the form of the standard normal density,
\[\nabla_{\!x} \phi(x-\mu) = -(x-\mu) \phi(x-\mu).\]
Partial integration over the intervals $]x_j, x_{j+1}[$ (where we let $x_0=-\infty$ and $x_{J+1} =\infty$) yields
\begin{align*}
E_\pi[(X - \mu)(m(X)-X)]
&= \int_{\mathbb{R}} \int_{\mathbb{R}} (x-\mu) \, (m(x)-x)\, \phi(x-\mu)\, dx\, d\pi(\mu)\\
&= -\sum_{j=0}^J \int_{\mathbb{R}} \int_{x_j}^{x_{j+1}}  (m(x)-x)\, \nabla_{\!x} \phi(x-\mu)\, dx\, d\pi(\mu)\\
&= \sum_{j=0}^J \int_{\mathbb{R}} \left [\int_{x_j}^{x_{j+1}}  (\nabla m(x)-1)\,  \phi(x-\mu)\, dx\right. \\
&+\left. \lim_{x\downarrow x_j}(m(x)-x) \phi(x-\mu) - \lim_{x\uparrow x_{j+1}} (m(x)-x) \phi(x-\mu)\right ] d\pi(\mu)\\
&=E_\pi[\nabla m(X)] - 1  + \sum_{j=1}^J \Delta m_j f(x_j).
\end{align*}
\hfill$\square$

\textbf{Proof of Lemma \ref{lemma:consr}:}
Uniform convergence of the first term follows by the exact same arguments we used to show uniform convergence of $L_n(\lambda)$ to $\bar R_\pi(\lambda)$ in Theorem \ref{theorem:l2conv}. We thus focus on the second term, and discuss its convergence on a case-by-case basis for our leading examples.

For ridge, this second term is equal to the constant
\[2\,\nabla_{\!x}m_R(x,\lambda) = \frac{2}{1+\lambda},\]
and uniform convergence holds trivially.

For lasso, the second term is equal to
\[2\, E_n[\nabla_{\!x}m_L(X,\lambda)] = 2\, P_n(|X| > \lambda).\]
To prove uniform convergence of this term we slightly modify the proof of the Glivenko-Cantelly Theorem (e.g., \cite{van1998asymptotic}, Theorem 19.1). Let $F_n$ be the cumulative distribution function of $X_1,\ldots, X_n$, and let $F_\pi$ be its population counterpart. It is enough to prove uniform convergence of $F_n(\lambda)$,
\[
\sup_{\pi\in\mathcal Q} P_\pi\left(\sup_{\lambda\in [0\infty]}\left|F_n(\lambda)-F_\pi(\lambda)\right|>\epsilon\right)\rightarrow 0\quad \forall\epsilon>0.
\]
Using Chebyshev's inequality and $\sup_{\pi\in\mathcal Q}\mbox{var}_\pi(1(X\leq \lambda))\leq 1/4$ for every $\lambda\in [0,\infty]$, we obtain
\[
\sup_{\pi\in\mathcal Q} |F_n(\lambda)-F_\pi(\lambda)|\stackrel{p}{\rightarrow} 0,
\]
for every $\lambda\in [0,\infty]$. Next, we will establish that for any $\epsilon>0$, it is possible to find a finite set of regularization parameters $0=\lambda_0<\lambda_1< \cdots <\lambda_k=\infty$ such that
\[
\sup_{\pi\in\mathcal Q} \max_{1\leq j\leq k}\{F_\pi(\lambda_j)- F_\pi(\lambda_{j-1})\}<\epsilon.
\]
This assertion follows from the fact that $f_\pi(x)$ is uniformly bounded by $\phi(0)$. The rest of the proof proceeds as in the proof of Theorem 19.1 in \cite{van1998asymptotic}.

Let us finally turn to pre-testing.
The objective function for pre-testing is equal to the one for lasso, plus additional terms for the jumps at $\pm \lambda$; the penalty term equals
\[2 P_n(|X| > \lambda) + 2 \lambda (\widehat{f}(-\lambda) + \widehat{f}(\lambda)).\]
Uniform convergence of the SURE criterion for pre-testing thus holds if (i) the conditions for lasso are satisfied, and (ii) we have a uniformly consistent estimator of $|x| \widehat{f}(x)$.\hfill$\square$

\textbf{Proof of Lemma \ref{lemma:cv}:} First, notice that the assumptions of the lemma plus convexity of the square function make $E_\pi[r_{n,k}(\lambda)]$ finite. Now, i.i.d.-ness of $(x_{1i},\ldots, x_{ki},\mu_i,\sigma_i)$ and mutual independence of $(x_1,\ldots, x_{k})$ conditional on $(\mu,\sigma^2)$ imply,
\begin{align*}
E_\pi[r_{n,k}(\lambda)] &= E_\pi\left [\left (m(X_{k-1},\lambda)- x_{k}\right )^2\right ]\\
&=E_\pi\left [\left (m(X_{k-1},\lambda)- \mu\right )^2\right ] +  E_\pi\left [\left (x_{k} - \mu\right )^2\right ]\\
&=\bar R_{\pi,k}(\lambda) + E_\pi[\sigma^2].
\end{align*}
\hfill$\square$

\textbf{Proof of Theorem \ref{theorem:cvunif}:}
We can decompose
\begin{align}
\label{equation:CVk}
r_{n,k}(\lambda)&= \frac{1}{n} \sum_{i=1}^n \left [\left (m(X_{k-1i},\lambda)- \mu_i\right )^2
+ (x_{ki} - \mu_i)^2 + 2 \left (m(X_{k-1i},\lambda)- \mu_i\right ) \left (x_{ki} - \mu_i\right )\right ]\nonumber\\
&= L_{n,k}(\lambda) + \frac{1}{n}\sum_{i=1}^n (x_{ki} - \mu_i)^2 - \frac{2}{n}\sum_{i=1}^n\left (m(X_{k-1i},\lambda)- \mu_i\right ) \left (x_{ki} - \mu_i\right ).
\end{align}
Theorem \ref{theorem:l2conv} and Lemma \ref{lemma:l2rlp} imply that the first term on the last line of equation (\ref{equation:CVk}) converges uniformly in quadratic mean to $\bar R_{\pi,k}(\lambda)$. The second term does not depend on $\lambda$. Uniform convergence in quadratic mean of this term to $-\bar v_\pi=E_\pi[\sigma_i^2]$ follows immediately from the assumption that $\sup_{\pi\in\mathcal Q} E_\pi[x_{k}^4]<\infty$. To prove uniform convergence to zero in quadratic mean of the third term, notice that,
\begin{align*}
E_\pi\Bigg[\Bigg(\frac{1}{n} \sum_{i=1}^n(m(X_{k-1i},\lambda)- \mu_i)  (x_{ki} - \mu_i)\Bigg)^2\Bigg]&=\frac{1}{n^2}\sum_{i=1}^n E_\pi\big[(m(X_{k-1i},\lambda)- \mu_i)^2  (x_{ki} - \mu_i)^2\big]\\
&\leq \frac{1}{n}\left(E_\pi\big[(m(X_{k-1},\lambda)- \mu)^4\big]  E_\pi\big[(x_{k} - \mu)^4\big]\right)^{1/2}\\
&\leq \frac{1}{n}\left(E_\pi\big[(X_{k-1}- \mu)^4+ \mu^4\big]  E_\pi\big[(x_{k} - \mu)^4\big]\right)^{1/2}.
\end{align*}
The condition $\sup_{\pi\in \mathcal Q} E[x_{j}^4]<\infty$ for $j=1,\ldots k$ guarantees that the two expectations on the last line of the last equation are uniformly bounded in $\pi\in\mathcal Q$, which yields the first result of the theorem.

The second result follows from Theorem \ref{theorem:loss}.
\hfill$\square$

\clearpage

\clearpage

\clearpage

\clearpage

\clearpage
\begin{table}
  \centering
  \caption{Average Compound Loss Across 1000 Simulations with $N=50$}
 
\hspace*{-2.1cm} \includegraphics[width = 1.25\textwidth]{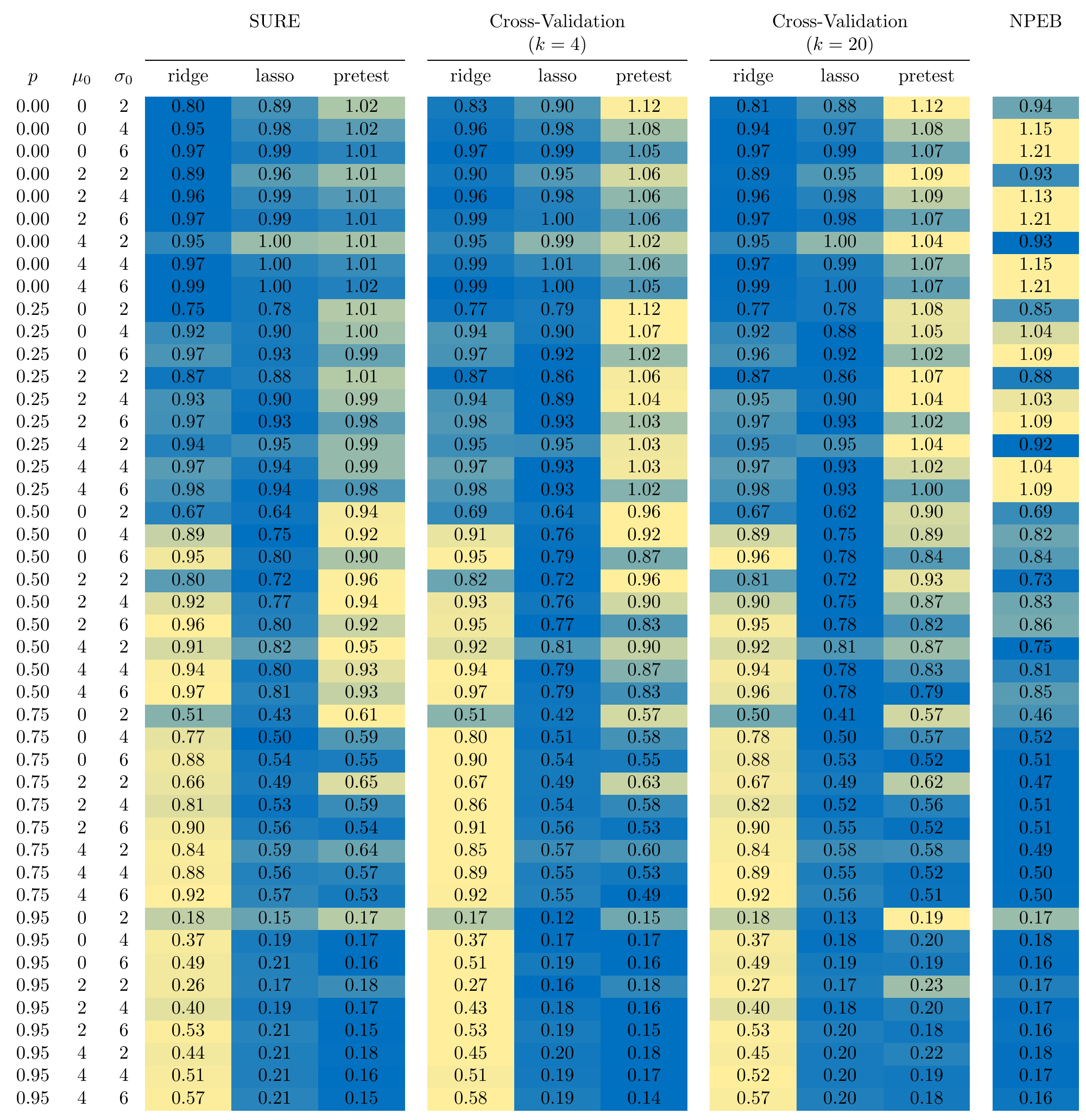}

  \label{table:simloss50}%
\end{table}%

\clearpage
\begin{table}
  \centering
  \caption{Average Compound Loss Across 1000 Simulations with $N=200$}
 
\hspace*{-2.1cm} \includegraphics[width = 1.25\textwidth]{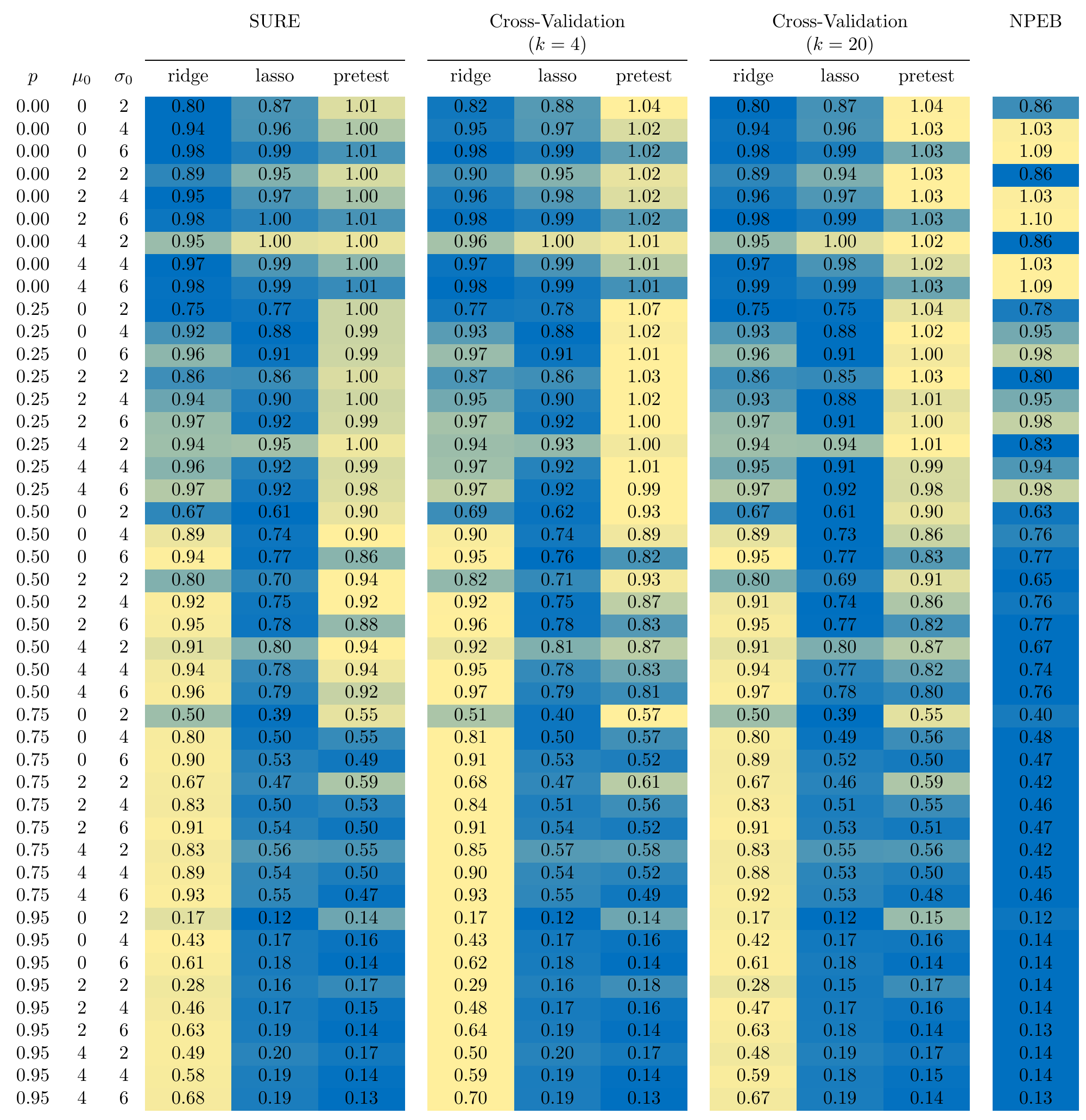}

  \label{table:simloss200}%
\end{table}%

\clearpage
\begin{table}
  \centering
  \caption{Average Compound Loss Across 1000 Simulations with $N=1000$}
 
\hspace*{-2.1cm} \includegraphics[width = 1.25\textwidth]{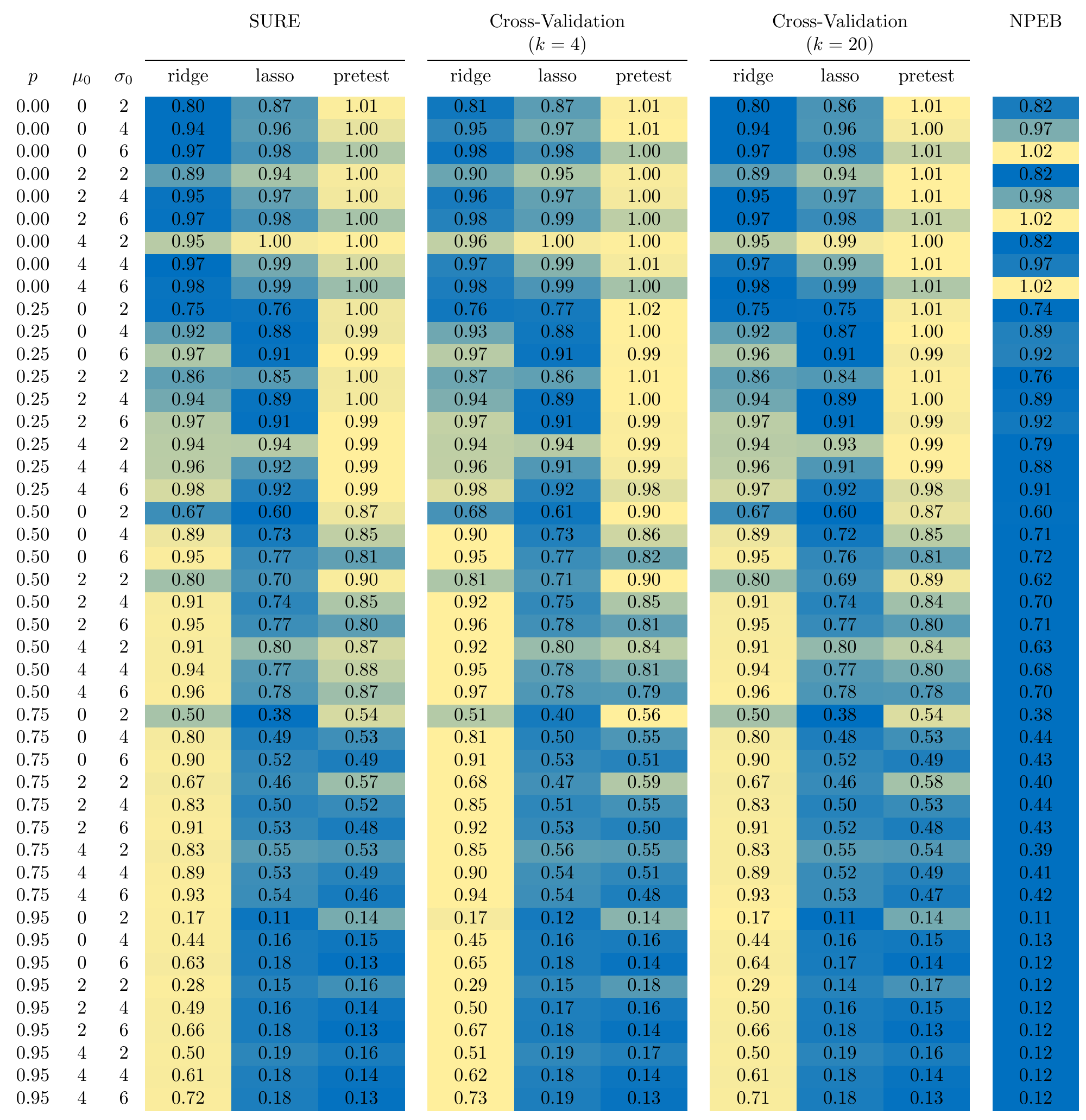}

  \label{table:simloss1000}%
\end{table}%

\clearpage
\bibliographystyle{chicago}
\bibliography{riskml_library}

\end{document}